\documentclass{article}

\usepackage[nonatbib,final]{neurips_2021}

\usepackage{enumitem}
\usepackage{wrapfig}
\usepackage[utf8]{inputenc} 
\usepackage[T1]{fontenc}    
\usepackage{hyperref}       
\usepackage{url}            
\usepackage{booktabs}       
\usepackage{amsfonts}       
\usepackage{nicefrac}       
\usepackage{microtype}      
\usepackage{xcolor}
\usepackage{amsmath}        
\usepackage{amssymb}
\usepackage{subfigure}
\usepackage{bm}
\usepackage{pdfpages}
\usepackage{microtype}
\usepackage{graphicx}
\usepackage{subfigure}
\usepackage{booktabs} 
\usepackage{hyperref}

\usepackage{color}
\usepackage[utf8]{inputenc} 
\usepackage{hyperref}       
\usepackage{url}            
\usepackage{nicefrac}       

\usepackage{amsthm,amsmath,amssymb,amsfonts,bbm}
\newtheorem{definition}{Definition}
\newtheorem{theorem}{Theorem}

\newtheorem{property}{Property}

\usepackage[ruled,vlined]{algorithm2e}
\usepackage{mathrsfs}
\usepackage{multirow}
\usepackage{bm}
\usepackage{enumitem}
\newcommand{\method}{FCFL}

\newcommand{\pan}[1]{{\color{black}{#1}}}

\title{Addressing Algorithmic Disparity and Performance Inconsistency in Federated Learning}

\author{Sen Cui\textsuperscript{1}\quad Weishen Pan\textsuperscript{1}\quad Jian Liang\textsuperscript{2}\quad Changshui Zhang\textsuperscript{1} \quad Fei Wang\textsuperscript{3}\\
\textsuperscript{1}Institute for Artificial Intelligence, Tsinghua University (THUAI),\\
State Key Lab of Intelligent Technologies and Systems,\\
Beijing National Research Center for Information Science and Technology (BNRist),\\
Department of Automation,Tsinghua University, Beijing, P.R.China\\
\textsuperscript{2} Alibaba Group, China\\
\textsuperscript{3} Department of Population Health Sciences, Weill Cornell Medicine, USA\\
{\tt\small \{cuis19, pws15\}@mails.tsinghua.edu.cn \quad xuelang.lj@alibaba-inc.com}\\
{\tt\small zcs@mail.tsinghua.edu.cn \quad few2001@med.cornell.edu}
}

\begin{document}

\maketitle

\begin{abstract}
Federated learning (FL) has gain growing interests for its capability of learning from distributed data sources collectively without the need of accessing the raw data samples across different sources. So far FL research has mostly focused on improving the performance, how the algorithmic disparity will be impacted for the model learned from FL and the impact of algorithmic disparity on the utility inconsistency are largely unexplored. In this paper, we propose an FL framework to jointly consider performance consistency and algorithmic fairness across different local clients (data sources). We derive our framework from a constrained multi-objective optimization perspective, in which we learn a model satisfying fairness constraints on all clients with consistent performance. Specifically, we treat the algorithm prediction loss at each local client as an objective and maximize the worst-performing client with fairness constraints through optimizing a surrogate maximum function with all objectives involved. A gradient-based procedure is employed to achieve the Pareto optimality of this optimization problem. Theoretical analysis is provided to prove that our method can converge to a Pareto solution that achieves the min-max performance with fairness constraints on all clients. Comprehensive experiments on synthetic and real-world datasets demonstrate the superiority that our approach over baselines and its effectiveness in achieving both fairness and consistency across all local clients.
\end{abstract}

\section{Introduction}

Federated learning (FL) \cite{yang2019federated} refers to the paradigm of learning from fragmented data without sacrificing privacy. FL has aroused broad interests from diverse disciplines including high-stakes scenarios such as loan approvals, criminal justice, healthcare, etc~\cite{xu2020federated}. An increasing concern is whether these FL systems induce disparity in local clients in these cases. For example, ProPublica reported that an algorithm used across the US for predicting a defendant’s risk of future crime produced higher scores to African-Americans than Caucasians on average~\cite{angwin2016machine}. This has caused severe concerns from the public on the real deployment of data mining models and made algorithmic fairness an important research theme in recent years.

Existing works on algorithmic fairness in machine learning have mostly focused on individual learning scenarios. There has not been much research on how FL will impact the model fairness at different local clients\footnote{We would like to emphasize that the algorithmic fairness we studied in this paper is not the federated fairness studied in \cite{li2019fair} and \cite{mohri2019agnostic}, as their fairness is referring to the performance at different clients, which is consistency.}. Recently, Du {\em et al.} \cite{du2020fairness} proposed a fairness-aware method, which considered the global fairness of the model learned through a kernel re-weighting mechanism. \pan{However, such a mechanism can not guarantee to achieve fairness at local clients in FL scenario, since different clients will have different distributions across protected groups.} For example, if we are building a mortality prediction model for COVID-19 patients within a hospital system \cite{vaid2021federated}, where each individual hospital can be viewed as a local client. Different hospitals will have different patient populations with distinct demographic compositions including race or gender. In this case, the model fairness at each hospital is important because that's where the model will be deployed, and it is unlikely that global model fairness can lead to local model fairness.


Due to the potential trade-off between algorithmic fairness and model utility, one aiming to mitigate the algorithmic disparity on local clients can exacerbate the inconsistency of the model performance (i.e., the model performance is different at different local clients). There have been researches \cite{li2019fair,mohri2019agnostic} trying to address the inconsistency without considering algorithmic fairness. In particular, Mohri {\em et al.} \cite{mohri2019agnostic} proposed an agnostic federated learning (AFL) algorithm that maximizes the performance of the worst performing client. Li {\em et al.} \cite{li2019fair} proposed a q-Fair Federated Learning (q-FFL) approach to weigh different local clients differently by taking the $q$-th power of the local empirical loss when constructing the optimization objective of the global model.



In this paper, we consider the problem of enforcing both algorithmic fairness and performance consistency across all local clients in FL. Specifically, suppose we have $N$ local clients, and $u_i$ represents the model utility for client $i$, and $g_i$ is the model disparity quantified by some computational fairness metric (e.g., demographic parity \cite{dwork2012fairness} or equal opportunity \cite{hardt2016equality}). Following the idea of AFL, we can maximize the utility of the worst-performed client to achieve performance consistency. We also propose to assign each client a "fairness budget" to ensure certain level of local fairness, i.e., $g_i\leq\epsilon_i(\forall i=1,2,\cdots,N)$ with $\epsilon_i$ being a pre-specified fairness budget for client $i$. Therefore, we can formulate our problem as a constrained multi-objective optimization framework as shown in Figure~\ref{fig:problem}, where each local model utility can be viewed as an optimization objective.

\pan{Since models with fairness and min-max performance may be not unique, we also require the model to be Pareto optimal. A model is Pareto optimal if and only if the utility of any client cannot be further optimized without degrading some others.} A Pareto optimal solution of this problem cannot be achieved by existing linear scalarization methods in federated learning (e.g., federated average, or FedAve in \cite{mcmahan2017communication}), as the non-i.i.d data distributions across different clients can cause a non-convex \textbf{Pareto Front} of utilities (all Pareto solutions form Pareto Front). Therefore, we propose \method, a new federated learning framework to obtain a fair and consistent model for all local clients. Specifically, we first utilize a surrogate maximum function (SMF) that considers the $N$ utilities involved simultaneously instead of the single worst, and then optimize the model to achieve Pareto optimality by controlling the gradient direction without hurting client utilities. Theoretical analysis proves that our method can converge to a fairness-constrained Pareto min-max model and experiments on both synthetic and real-world data sets show that \method~can achieve a Pareto min-max utility distribution with fairness guarantees in each client. The source codes of \method~ are made publicly available at \url{https://github.com/cuis15/FCFL}.




\section{Related Work}

Algorithm fairness is defined as the disparities in algorithm decisions made across groups formed by protected variables (such as gender and race). Some approaches have been proposed to mathematically define if an algorithm is fair. For example, demographic parity \cite{dwork2012fairness} requires the classification results to be independent of the group memberships and equalized opportunity \cite{hardt2016equality} seeks for equal false negative rates across different groups. Besides, there are also recent works focusing on the interpretation of the source of model disparity~\cite{pan2021explaining}. Plenty of approaches have been proposed to reduce model disparity. One type of method is to train a classifier first and then post-adjust the prediction by setting different thresholds for different groups~\cite{hardt2016equality} or learning a transformation function~\cite{cui2021towards,kallus2019fairness}. Other methods have been developed for optimization of fairness metrics during the model training process through adversarial learning~\cite{beutel2017data,louizos2015variational,madras2018learning,zemel2013learning,zhang2018mitigating} or regularization~\cite{kamishima2011fairness,zafar2015fairness,beutel2019putting}. Du {\em et al.} \cite{du2020fairness} considered algorithm fairness in the federated learning setting and proposed a regularization method that assigns a reweighing value on each training sample for loss objective and fairness regularization to deal with the global disparity. This method cannot account for the discrepancies among the model disparities at different local clients. In this paper, we propose to treat fairness as a constraint and optimize a multi-objective optimization with multiple fairness constraints for all clients while maximally maintain the model utility.

As we introduced in the introduction, the global consensus model learned from federated learning may have different performances on different clients. There are existing research trying to address such inconsistency issue by maximizing the utility of the worst-performing client. In particular, Li {\em et al.} \cite{li2019fair} propose q-FFL to obtain a min-max performance of all clients by empirically adjusting the power of the objectives, which cannot always guarantee a more consistent model utility distribution without sufficient searching for appropriate power values. Mohri {\em et al.} \cite{mohri2019agnostic} propose AFL, a min-max optimization scheme which focuses on the single worst client. However, focusing on the single worst objective can cause another client to perform worse, thus we propose to take all objectives into account and optimize a surrogate maximum function to achieve a min-max performance distribution in this paper.

Multi-objective optimization aims to learn a model that gives consideration to all objectives involved. The optimization methods for multi-objective typically involve linear scalarization or its variants, such as those with adaptive weights \cite{chen2018gradnorm}, but it is challenging for these approaches to handling the competing performance among different clients \cite{mahapatra2020multi}. Martinez {\em et al.} \cite{martinez2020minimax} proposed a multi-objective optimization framework called Min-Max Pareto Fairness (MMPF) to achieve consistency by inducing a min-max performance of all groups based on convex assumption, which is fairly strong as non-convex objectives are ubiquitous. In this paper, we formulate the problem of achieving both fairness and consistency in federated networks through constrained multi-objective optimization. Previous research on solving this problem has been mostly focusing on gradient-free algorithms such as evolutionary algorithms \cite{coello2006evolutionary}, physics-based and deterministic approaches \cite{evtushenko2014deterministic}. Gradient-based methods are still under-explored \cite{zerbinati2011comparison}. We propose a novel gradient-based method \method~, which searches for the desired gradient direction iteratively by solving constrained Linear Programming (LP) problems to achieve fairness and consistency simultaneously in federated networks.

\section{Problem Setup}

The problem to be solved in this paper is formally defined in this section. Specifically, we will introduce the algorithmic fairness problem, how to extend existing fairness criteria to federated setting, and the consistency issues of model utility in federated learning.

\subsection{Preliminaries}

\noindent \textbf{Federated Learning.}
Suppose there are $N$ local clients $c_1, c_2, ... c_N$ and each client is associated with a specific dataset $\mathcal{D}^{k} = \left\{X^{k}, Y^{k}\right\}$, $k \in \left\{1,2,...,N \right\}$, where the input space $X^{k}$ and output space $Y^{k}$ are shared across all $N$ clients. There are $n^{k}$ samples in the $k$-th client and each sample is denoted as $\left\{ x^{k}_{i}, y^{k}_{i} \right\}$. The goal of the federated learning is to collaboratively learn a \pan{global} model $h$ with the parameters $\theta$ to predict the label $\hat{Y}^{k}$ as on each client. The classical federated learning aims to minimize the empirical risk over the samples from all clients i.e., $\min_{\theta} \quad \frac{1}{\sum_{k=1}^{N} n^{k}} \sum\nolimits_{k=1}^{N} \sum\nolimits_{i=1}^{n^{k}} l_{k}(h_{\theta}(x^{k}_i), y^{k}_{i})$ where $l_{k}$ is the loss objective of the $k$-th client.

\noindent \textbf{Fairness.}
Fairness refers to the disparities in the algorithm decisions made across different groups formed by the sensitive attribute, such as gender and race. If we denote the dataset on the $k$-th client as $D^{k} = \left\{ X^{k}, A^{k}, Y^{k} \right\}$, where $A^{k} \in \mathcal{A}$ is the binary sensitive attribute, then we can define the multi-client fairness as follows:
\begin{definition}[Multi-client fairness (MCF)]A learned model $h$ achieves multi-client fairness if $h$ meets the following condition:
\end{definition}
\begin{equation}
  \begin{aligned}
    \Delta Dis_{k} (h)- \epsilon_{k} \leq 0 \quad \forall k \in \left\{1,.,N\right\}
  \end{aligned}
  \label{eq:faircriteria_FL}
\end{equation}
\label{def:1}
where $\Delta Dis_{k}(h)$ denotes the disparity induced by the model $h$ and $\epsilon_k$ is the given fairness budget of the $k$-th client. The disparity on the $k$-th client $\Delta Dis_{k}$ can be measured by \emph{demographic parity} (DP) \cite{dwork2012fairness} and \emph{Equal Opportunity} (EO) \cite{hardt2016equality}  as follows:
\begin{equation}
  \begin{aligned}
    & \Delta DP_{k} = | P(\hat{Y}^{k}=1 | A^{k} = 0) - P(\hat{Y}^{k}=1 | A^{k} = 1)|\\
    & \Delta EO_{k} = | P(\hat{Y}^{k}=1 | A^{k} = 0, Y^{k} = 1) - P(\hat{Y}^{k}=1 | A^{k} = 1, Y^{k} = 1)|
  \end{aligned}
  \label{eq:DP_EO}
\end{equation}
As data heterogeneity may cause different disparities across all clients, the fairness budgets $\epsilon_k$ in Definition \ref{def:1} specifies the tolerance of model disparity at the $k$-th client.

\noindent \textbf{Consistency.}
Due to the discrepancies among data distributions across different clients, the model performance on different clients could be different. Morever, the inconsistency will be magnified when we adjust the model to be fair on local clients. There are existing research trying to improve the model consistency by maximizing the utility of the worst performing client \cite{li2019fair,mohri2019agnostic}: $$\min_{\theta} \ \max_{k \in \left\{1,.,N\right\} } \frac{1}{n^{k}} \sum\nolimits_{i=1}^{n^{k}} l_k(h_{\theta}(x^{k}_{i}), y^{k}_{i})$$ where the $\max$ is over the losses across all clients.

\subsection{Fair and Consistent Federated Learning (\method)}
\begin{figure}[h!]
  \setlength{\abovecaptionskip}{-0.1cm}
  \setlength{\belowcaptionskip}{-0.2cm}
  \centering
  \subfigure[]{
    \centering
    \includegraphics[width=.39\columnwidth]{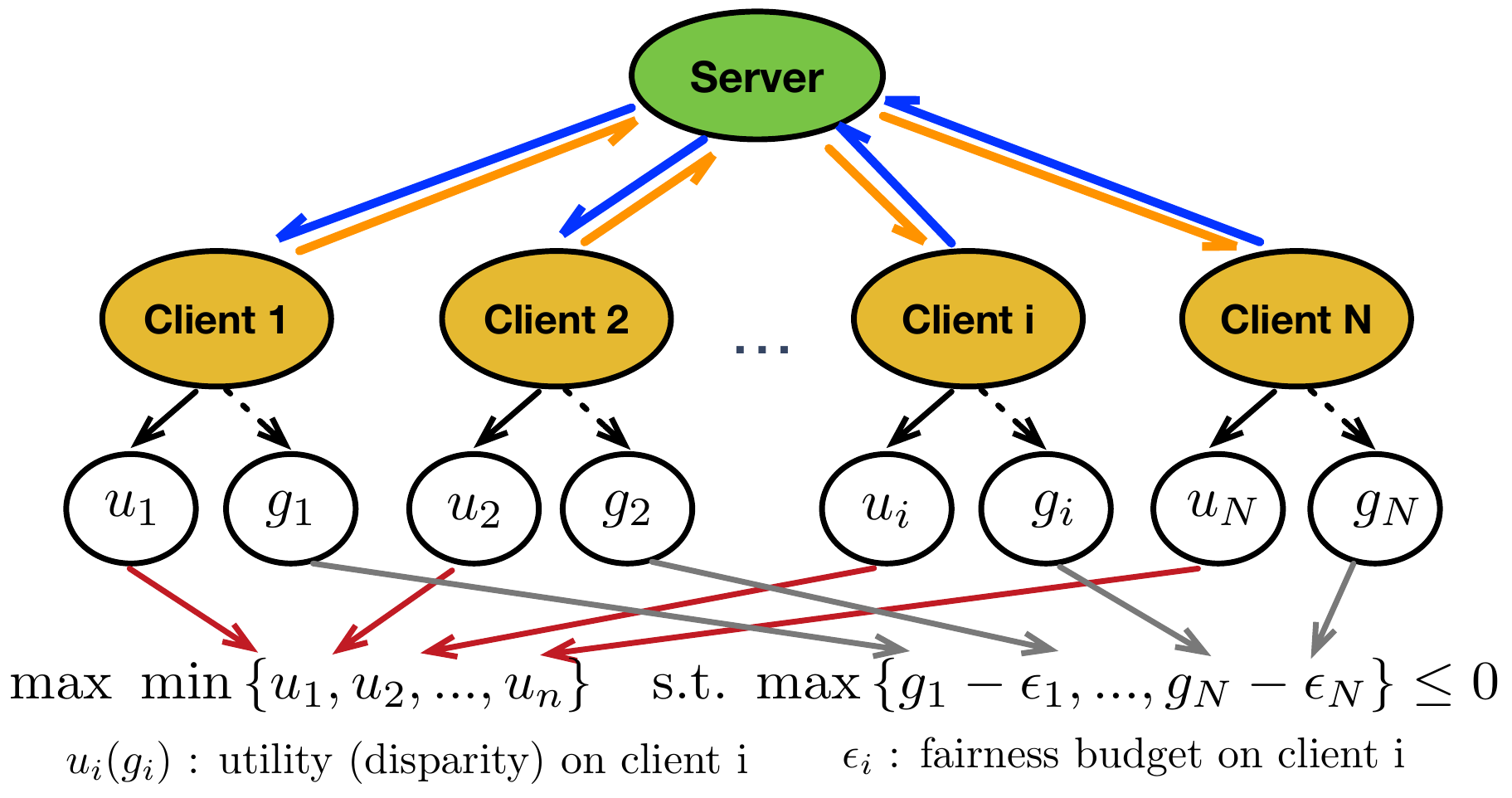}
    \label{fig:problem}
  }%
  \subfigure[]{
    \centering
    \includegraphics[width=0.27\columnwidth]{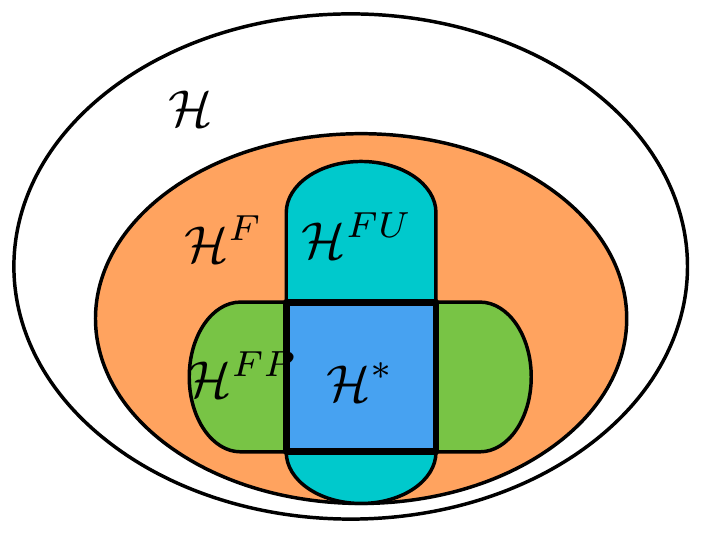}
    \label{fig:desired_PT}
  }%
  \subfigure[]{
    \centering
    \includegraphics[width=0.27\columnwidth]{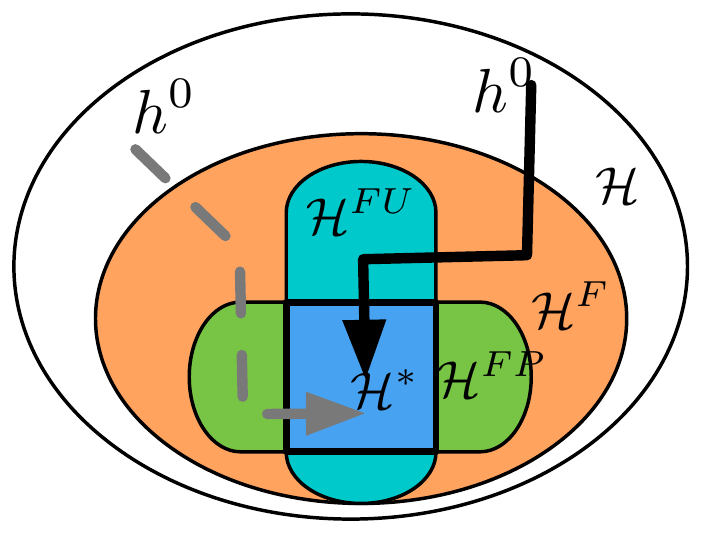}
    \label{fig:optimization_path}
  }%
  \caption{(a) The architecture of our proposed fairness-constrained min-max problem. (b) The relationship of the 5 hypothesis sets involved and $\mathcal{H}^{*} =  \mathcal{H}^{FP} \cap \mathcal{H}^{FU}$ is the desired hypothesis set. (c) Two optimization paths to achieve a fair Pareto min-max model: 1) $h^{0} \rightarrow \mathcal{H}^{FP} \rightarrow \mathcal{H}^{*}$: the gray dotted line represents that the initial model $h^{0}$ first achieves Pareto optimality with MCF then achieves min-max performance; 2) $h^{0} \rightarrow \mathcal{H}^{FU} \rightarrow \mathcal{H}^{*}$: the black solid line denotes that the initial model $h^{0}$ first achieves min-max performance with MCF then achieves Pareto optimality.}
  \label{fig:sets}
\end{figure}

Our goal is to learn a model $h$ which 1) satisfies MCF as we defined in Definition \ref{def:1}; 2) maintains consistent performances across all clients. We will use $\Delta DP_{k}$ defined in Eq.(\ref{eq:DP_EO}) as measurement of disparity in our main text while the same derivations can be developed when adapting other metrics, so we have $g_k(h(X^{k}), A^{k}) = \Delta DP_{k},$ and $g_k$ is the function of calculating model disparity on the $k$-th client. Similarly, the model utility loss $l_{k}(h(X^{k}), Y^{k})$ can be evaluated by different metrics (such as cross-entropy, hinge loss and squared loss, etc). In the rest of this paper we will use $l_{k}(h)$ ($g_{k}(h)$) for $l_{k}(h(X^{k}), Y^{k})$ ($g_{k}(h(X^{k}), A^{k})$) without causing further confusions. We formulate \method as the problem of optimizing the $N$ utility-related objectives $\left\{ l_{1}(h), l_{2}(h), ... ,l_{N}(h) \right\}$ to achieve \textbf{Pareto Min-Max} performance with $N$ fairness constraints:


\begin{equation}
  \begin{aligned}
    \min_{h \in \mathcal{H}} \quad [ l_{1}(h), l_{2}(h), ... l_{N}(h)] \quad \text{s.t.} \ g_{k}(h) - \epsilon_{k} \leq 0 \quad \forall k \in \left\{1,.,N\right\}.
  \end{aligned}
  \label{eq:obj1}
\end{equation}
The definitions of \emph{Pareto Solution} and \emph{Pareto Front}, which are fundamental concepts in multi-objective optimization, are as follows:
\begin{definition}[Pareto Solution and Pareto Front] Suppose $\bm{l(h)} = [l_{1}(h), l_{2}(h), ... l_{N}(h)]$ represents the utility loss vector on $N$ learning tasks with hypothesis $h \in \mathcal{H}$, we say $h$ is a Pareto Solution if there is no hypothesis $h'$ that dominates $h$: $h' \prec h$, i.e., $$ \nexists h' \in \mathcal{H}, \ \text{s.t.} \ \forall i : l_{i}(h') \leq l_{i}(h) \ \text{and} \ \exists j: \ l_{j}(h') < l_{j}(h).$$  All Pareto solutions form Pareto Front $\mathcal{P}$.
  \label{def:2}
\end{definition}
From Definition \ref{def:2}, for a given hypothesis set $\mathcal{H}$ and the objective vector $\bm{l(h)}$, the Pareto solution avoids unnecessary harm to client utilities and may not be unique. We prefer a Pareto solution that achieves a higher consistency. Following the work in \cite{li2019fair,mohri2019agnostic}, we want to obtain a Pareto solution $h^{*}$ with min-max performance. Figure \ref{fig:desired_PT} shows the relationships among different model hypothesis sets, and we explain the meanings of different notations therein as follows:

\textbf{(1)} $\mathcal{H}^{F}$ is the set of model hypotheses satisfying MCF, i.e.,  $$g_{k}(h) -\epsilon_{k} \leq 0 \quad \forall k \in \left\{1,.,N\right\}, \ \forall h \in \mathcal{H}^{F}.$$
\textbf{(2)} $\mathcal{H}^{FU}$ is the set of model hypotheses achieving min-max performance (consistency) with MCF:
  \begin{equation}
    \begin{aligned}
      \mathcal{H}^{FU} \subset \mathcal{H}^{F} \quad \text{and} \quad l_{max}(h') \leq l_{max}(h), \ \forall h \in \mathcal{H}^{F}, \ \forall h' \in \mathcal{H}^{FU}.
    \end{aligned}
    \label{eq:obj4_0}
  \end{equation}
\textbf{(3)} $\mathcal{H}^{FP}$ is the set of model hypotheses achieving Pareto optimality with MCF:
  \begin{subequations}
    \begin{align}
      & \mathcal{H}^{FP} \subset \mathcal{H}^{F} \label{eq:obj21_2} \\
      & h' \nprec h, \ \forall h' \in \mathcal{H}^{F}, \ \forall h \in \mathcal{H}^{FP}  \label{eq:obj21_3}
    \end{align}
    \label{eq:obj2}
  \end{subequations}
where Eq.(\ref{eq:obj21_2}) satisfies $\forall h \in \mathcal{H}^{FP}$ meets MCF, and Eq.(\ref{eq:obj21_3}) ensures that $\forall h \in \mathcal{H}^{FP}$ is a Pareto model with MCF.

\textbf{(4)} $\mathcal{H}^{*}$ is our desired set of model hypotheses achieving Pareto optimality and min-max performance with MCF: $\mathcal{H}^{*} =  \mathcal{H}^{FP} \cap \mathcal{H}^{FU}$.





\begin{theorem}
(proof in Appendix) The hypothesis set $\mathcal{H}^{*}$ is non-empty, i.e.,
\begin{equation}
\exists h \in \mathcal{H}, h \in \mathcal{H}^{*}.
\end{equation}
\label{theorem:1}
\vspace{-.3cm}
\end{theorem}
In summary, from Theorem~\ref{theorem:1}, our goal is to obtain a fair and consistent model $h^{*} \in \mathcal{H}^{*}$ to achieve Pareto optimality and min-max performance with MCF.
\section{Fairness-Constrained Min-Max Pareto Optimization}

\subsection{Preliminary: Gradient-based Multi-Objective Optimization}
Pareto solutions of the multi-objective optimization problem can be reached by gradient descent procedures. Specifically, given the initial hypothesis $h$ with parameter $\theta^{0} \in \mathbb{R}^{n}$, we optimize $h_{\theta^{t}}$ by moving to a specific gradient direction $d$ with a step size $\eta$: $\theta^{t+1} = \theta^{t} + \eta d$. $d$ is a descent direction if it decreases the objectives ($l_{i}(h_{\theta^{t+1}}) \leq l_{i}(h_{\theta^t}), \ \forall i$).
Suppose $\nabla_{\theta} l_{i}$ is the gradient of the $i$-th objective $l_{i}$ with respect to the parameter $\theta$, if we select $d^{*}$ which satisfies $d^{*T} \nabla_{\theta} l_{i} \leq 0$ for all $i$, $d^{*}$ is a descent direction and $l_{i}(h_{\theta^{t}})$ decreases after the $t+1$ iteration.

If we directly search for the descent direction $d$ to achieve $d^{T} \nabla_{\theta} l_{i} \leq 0, \forall i$, the computation cost can be tremendous when $d$ is a high-dimensional vector. D{\'e}sid{\'e}ri \emph{et al.} \cite{desideri2012multiple} proposed to find a descent direction $d$ in the convex hull of the gradients of all objectives denoted as $G  = [\nabla_{\theta} l_{1}, \nabla_{\theta} l_{2}, ... \nabla_{\theta} l_{N}]$ by searching for a $N$-dimension vector $\alpha^{*}$ (typically $N \ll n$ in deep model), which is formulated as follows:
\begin{equation}
  \begin{aligned}
    d = \alpha^{*T} G  \in \overline{G} \quad  \text{where} \ \overline{G} = \left \{ \sum\nolimits_{i=1}^{N} \alpha_{i} \nabla_{\theta}l_{i} \mid \alpha_{j} \geq 0 \ \forall j \ \text{and} \ \sum\nolimits_{j=1}^{N} \alpha_{j} = 1 \right\}.
    \label{eq:d_com}
  \end{aligned}
\end{equation}

\subsection{Overview of the Optimization Framework}

To obtain a \textbf{fair Pareto min-max} model, there are two optimization paths shown in Figure \ref{fig:optimization_path}. The gray dotted line denotes the optimization path where we first achieve Pareto optimality with MCF, then we try to achieve min-max performance while keeping Pareto optimality. However, it's hard to adjust a Pareto model to another~\cite{mahapatra2020multi}. Therefore, we propose to first achieve min-max performance with MCF then achieve Pareto optimality as the black solid line in Figure \ref{fig:optimization_path}. In particular, we propose a two-stage optimization method for this constrained multi-objective problem: 1) constrained min-max optimization to achieve a fair min-max model; 2) constrained Pareto optimization to continue to optimize the model to achieve Pareto optimality while keeping min-max performance with MCF.

\noindent \textbf{Constrained Min-Max Optimization}
We define a constrained min-max optimization problem on the hypothesis set $\mathcal{H}$:
\begin{equation}
  \begin{aligned}
    & \min_{h \in \mathcal{H}} \quad l_{max}(h)\\
    & \quad \quad l_{max}(h) = \max \left(l_{i}(h) \right) \quad i \in \left\{1,.,N\right\} \\
    & \text{s.t.} \quad g'_{max}(h) \leq 0 \\
    & \quad \quad g'_{max}(h) = \max \left( g'_{i}(h)\right) = \max \left( g_{i}(h) - \epsilon_{i} \right)  \quad  i \in \left\{1,.,N\right\}
  \end{aligned}
  \label{eq:obj5_1}
\end{equation}
where $\epsilon_{i}$ is the given fairness budget on the $i$-th client. By optimizing the constrained objective in Eq.(\ref{eq:obj5_1}), we obtain a model $h^{1} \in \mathcal{H}^{FU}$ that 1) $h^{1}$ satisfies MCF; 2) $h^{1}$ achieves the optimal utility on the worst performing client.

\noindent \textbf{Constrained Pareto Optimization}
Caring only about the utility of the worst-performing client can lead to unnecessary harm to other clients since the rest objectives can be further optimized. Therefore, we then continue to optimize $h^{1} \in \mathcal{H}^{FU}$ to achieve Pareto optimality:
\begin{subequations}
  \begin{align}
    & \min_{h \in \mathcal{H}} \frac{1}{N} \sum\nolimits_{i=1}^{N} l_{i}(h) \label{eq:obj4_2_1}\\
    & \text{s.t.} \ l_{i}(h) \leq l_{i}(h^{1}) \quad  \forall {i \in \left\{1,.,N\right\}} \label{eq:obj4_2_2}\\
    & \quad \quad g'_{max}(h) \leq 0 \label{eq:obj4_2_3}\\
    & \quad \quad g'_{max}(h) = \max \left( g'_{i}(h)\right) = \max \left( g_{i}(h) - \epsilon_{i} \right)  \quad  i \in \left\{1,.,N\right\} \label{eq:obj4_2_4}
  \end{align}
  \label{eq:obj4_2}
\end{subequations}
where we optimize $h$ without hurting the model utility on any client so that the converged Pareto model $h^{*}$ of Eq.(\ref{eq:obj4_2}) satisfies $l_{max}(h^{*}) \leq l_{max}(h^{1})$. Morever, $h^{*}$ satisfies MCF as the constraint in Eq.(\ref{eq:obj4_2_3}), so $h^{*} \in \mathcal{H}^{*}$ is a Pareto min-max model with $N$ fairness constraints on all clients.

\subsection{Achieving Min-Max Performance with MCF}
\label{sec:4.3}

Minimizing the current maximum value of all objectives directly in Eq.(\ref{eq:obj5_1}) can cause another worse objective and can be computationally hard when faced with a tremendous amount of clients. We will use $l_{m}(h)$ ($g'_{m}(h)$) to denote $l_{max}(h)$ ($g'_{max}(h)$) without causing further confusions for expression simplicity. We propose to use a smooth surrogate maximum function (SMF) \cite{polak2003algorithms} to approximate an upper bound of $l_{m}(h)$ and $g_{m}(h)$ as follows:
\begin{equation}
  \begin{aligned}
  \hat{l}_{m}(h, \delta_{l})= \delta_{l} \ln \sum\nolimits_{i=1}^{N} \exp \left(\frac{l_{i}(h)}{\delta_{l}}\right), \ \hat{g'}_{m}(h, \delta_{g})= \delta_{g} \ln \sum\nolimits_{i=1}^{N} \exp \left(\frac{g'_{i}(h)}{\delta_{g}}\right), \ (\delta_{g}, \delta_{l}>0).
  \end{aligned}
  \label{eq:appro_obj1}
\end{equation}
It is obvious that $l_{m}(h) \leq \hat{l}_{m}(h, \delta_{l})  \leq  l_{m}(h) + \delta_{m} \ln(N)$ and $\lim_{\delta_{l} \to 0^{+}} \hat{l}_{m}(h(x), \delta_{l}) = l_{m}(h)$. For $\hat{g'}_{m}(h, \delta_{g})$, we can get a similar conclusion. The objective in Eq.(\ref{eq:obj5_1}) is approximated as follows:
\begin{subequations}
  \begin{align}
    & \min_{h \in \mathcal{H}} \quad \hat{l}_{m}(h, \delta_{l}) \label{eq:obj5_2_1}\\
    & \text{s.t.} \quad \hat{g'}_{m}(h, \delta_{g}) \leq 0 \label{eq:obj5_2_2}.
  \end{align}
  \label{eq:obj5_2}
\end{subequations}
\begin{property}
  \pan{There always exists an initial trivial model which satisfies the MCF criterion by treating all samples equally (e.g., $h(x) = 1, \forall x$).}
  \label{prop:4}
\end{property}



From Property \ref{prop:4}, we can always initialize $h$ to satisfy $\hat{g'}_{m}(h, \delta_{g}) \leq 0$ in Eq.(\ref{eq:obj5_2_2}). Then we optimize the upper bound of $l_{m}(h)$ when ensuring MCF. As the hypothesis $h$ owns the parameter $\theta$, we use $\nabla_{\theta}\hat{l}$ and $\nabla_{\theta}\hat{g'}$ to represent the gradient $\nabla_{{\theta}}\hat{l}_{m}(h_{\theta}, \delta_{l})$ and $\nabla_{{\theta}}\hat{g'}_{m}(h_{\theta}, \delta_{g})$, respectively. We propose to search for a descent direction $d$ for Eq.(\ref{eq:obj5_2}) in the convex hull of $G = [\nabla_{\theta} \hat{l}, \nabla_{\theta} \hat{g'}]$ in two cases where $d$ is defined in Eq.(\ref{eq:d_com}). For the $t$-th iteration:\\
\textbf{(1)} if $h_{\theta^{t}}$ satisfies the fairness constraint defined in Eq.(\ref{eq:obj5_2_2}), we focus on the descent of $\hat{l}_{m}(h_{\theta^{t}}, \delta_{l}^{t})$:
  \begin{equation}
    \min_{d \in \overline{G}} \quad  d^{T} \nabla_{\theta^{t}} \hat{l} \quad \text{if} \ \hat{g'}_{m}(h_{\theta^{t}}, \delta_{g}^{t}) \leq 0,
    \label{eq:obj5_3_1}
  \end{equation}
\textbf{(2)} if $h_{\theta^{t}}$ violates the fairness constraint, we aim to the descent of $\hat{g'}_{m}(h_{\theta^{t}}, \delta_{g}^{t})$ without causing the ascent of $\hat{l}_{m}(h_{\theta^{t}}, \delta_{l}^{t})$ :
  \begin{subequations}
    \begin{align}
      \min_{d \in \overline{G}} \quad  d^{T} \nabla_{\theta^{t}} \hat{g'},  \quad \text{s.t.} \ d^{T} \nabla_{\theta^{t}} \hat{l} \leq 0.
    \end{align}
    \label{eq:obj5_3_2}
    \vspace{-.2cm}
  \end{subequations}
If the obtained gradient $d$ satisfies $||d||^{2} \leq \epsilon_{d}$, we decrease the parameter $\delta^{t}_{l}, \delta^{t}_{g}$ as:
\begin{equation}
  \begin{aligned}
    & \delta_{l}^{t+1} = \beta \cdot \delta_{l}^{t}, \ \delta_{g}^{t+1} = \beta \cdot \delta_{g}^{t} \quad & \text{if}\ ||d||^{2} \leq \epsilon_{d}\\
    & \delta_{l}^{t+1} = \delta_{l}^{t}, \ \delta_{g}^{t+1} = \delta_{g}^{t} \quad & \text{otherwise},
  \end{aligned}
  \label{eq:obj5_3_3}
\end{equation}
where $\beta$ is the decay ratio that $0 < \beta < 1$. From Eq.(\ref{eq:obj5_3_3}), we narrow the gap between $\hat{l}_{m}(h, \delta_{l})$ and $l_{max}(h)$ by decreasing the parameter $\delta_{l}$ as every time the algorithm approaches convergence. From Eq.(\ref{eq:obj5_3_1}) and Eq.(\ref{eq:obj5_3_2}), we optimize either $\hat{l}_{m}$ or $\hat{g'}_{m}$ and keep $\hat{l}_{m}$ without ascent in each iteration.

\subsection{Achieving Pareto Optimality \emph{and} Min-Max Performance with MCF}
\label{sec:4.4}
As the model $h^{1}_{\theta} \in \mathcal{H}^{FU}$ obtained from constrained min-max optimization cannot guarantee Pareto optimality, we continue to optimize $h_{\theta}^{1}$ to be a Pareto model without causing the utility descent on any client.
We propose a constrained linear scalarization objective to reach a Pareto model and the $t$-th iteration is formulated as follows:
\begin{subequations}
  \begin{align}
    & \min_{d \in \overline{G}}  \frac{1}{N}\sum_{i=1}^{N}  d^{T} \nabla_{{\theta}^{t}} l_{i} \label{eq:obj5_4_1}\\
    & \text{s.t.} \quad d^{T} \nabla_{{\theta}^{t}} l_{i} \leq 0 \quad \forall i \in \left\{1,.,N\right\} \label{eq:obj5_4_2} \\
    & \quad \quad d^{T} \nabla_{{\theta}^{t}} \hat{g'}_{m} \leq 0 \label{eq:obj5_4_3},
  \end{align}
  \label{eq:obj5_4}
\end{subequations}
where $G = [\nabla_{\theta} l_{1}, \nabla_{\theta} l_{2}, ... \nabla_{\theta} l_{N}, \nabla_{\theta} g_{1}, \nabla_{\theta} g_{2}, ..., \nabla_{\theta} g_{N}]$ and $\overline{G}$ is the convex hull of $G$. The non-positive angle of $d$ with each gradient $\nabla_{{\theta}^{t}} l_{i}$ ensures that all objective values decrease. Similarly, if we aim to reach the Pareto solution without causing utility descent only on the worst performing client, the constraint in Eq.(\ref{eq:obj5_4_2}) is replaced by $d^{T} \nabla_{{\theta}^{t}} \hat{l} \leq 0$.

Different from constrained min-max optimization in Section \ref{sec:4.3} where the objective to be optimized in each iteration depends on whether $\hat{g}_{m}(h_{\theta^{t}}, \delta^{t}_{g}) \leq 0$, in constrained Pareto optimization procedure, as we have achieved MCF, we optimize the objective in Eq.(\ref{eq:obj5_4}) for a dominate model in each iteration. Specifically, we restrict $\hat{g'}_{m} \leq 0$ to keep fairness on each client given the reached hypothesis $h^{1} \in \mathcal{H}^{FU}$. Meanwhile, we constrain $l_{i}(h)$ to descend or remain unchanged until any objective cannot be further minimized. Algorithm 1 in Appendix shows all steps of our method. The convergence analysis and the time complexity analysis of our framework are in Appendix. Moreover, we also present our discussion about the fairness budget $\epsilon$ in Appendix.

\section{Experiments}
We intuitively demonstrate the behavior of our method by conducting experiments on synthetic data. For the experiments on two real-world federated datasets with fairness issues, we select two different settings to verify the effectiveness of our method: (1) assign equal fairness budgets for all local clients; (2) assign client-specific fairness budgets. More experimental results and detailed implementation are in Appendix.
\subsection{Experimental Setup}
\noindent \textbf{Federated Datasets} (1) Synthetic dataset: following the setting in \cite{lin2019pareto,mahapatra2020multi}, the synthetic data is from two given non-convex objectives; (2) UCI \emph{Adult} dataset \cite{mohri2019agnostic}: \emph{Adult} contains more than 40000 adult records and the task is to predict whether an individual earns more than 50K/year given other features. Following the federated setting in \cite{li2019fair,mohri2019agnostic}, we split the dataset into two clients. One is PhD client in which all individuals are PhDs and the other is non-PhD client. In our experiments, we select \emph{race} and \emph{gender} as sensitive attributes, respectively. (3)\emph{eICU} dataset: We select  \cite{pollard2018eicu}, a clinical dataset collecting patients about their admissions to ICUs with hospital information. Each instance is a specific ICU stay. We follow the data preprocessing procedure in \cite{johnson2018generalizability} and naturally treat 11 hospitals as 11 local clients in federated networks. We conduct the task of predicting the prolonged length of stay (whether the ICU stay is longer than 1 week, ) and select $race$ as the sensitive attribute.


\textbf{Evaluation Metrics} (1) Utility metric: we use $accuracy$ to measure the model utility in our experiments; (2) Disparity metrics: our method is compatible with various of fairness metrics. In our experiments, we select two metrics (marginal-based metric \emph{Demographic Parity} \cite{dwork2012fairness} and conditional-based metric \emph{Equal Opportunity}  \cite{hardt2016equality} (The results of \emph{Equal Opportunity} are in the Appendix) to measure the disparities defined in Eq.(\ref{eq:DP_EO}); (3) Consistency: following the work \cite{li2019fair,mohri2019agnostic}, we use the utility on the worst-performing client to measure consistency.

\textbf{Baselines} As we do not find prior works proposed for achieving fairness in each client, we select FA \cite{du2020fairness}, MMPF \cite{martinez2020minimax} and build FedAve+FairReg as baselines in our experiments. For all baselines, we try to train the models to achieve the optimal utility with fairness constraints. If the model cannot satisfy the fairness constraints, we keep the minimum of disparities with reasonable utilities. (1) MMPF \cite{martinez2020minimax}, Martinez {\em et al.} develop MMPF which optimizes all objectives on convex assumption to induce a min-max utility of all groups; (2) FA \cite{du2020fairness}, Du {\em et al.} propose FA, a kernel-based model-agnostic method with regularizations for addressing fairness problem on a new unknown client instead of all involved clients; (3) FedAve+FairReg, we build FedAve+FairReg, which optimizes the linear scalarized objective with the fairness regularizations of all clients.

\begin{wrapfigure}{l}{8cm}
  \vspace{-.5cm}
  \setlength{\abovecaptionskip}{0.2cm}
  \setlength{\belowcaptionskip}{-0.2cm}
  \centering
  \subfigure[$\theta^{0}$ violates the constraints]{
    \centering
    \includegraphics[width=0.25\columnwidth]{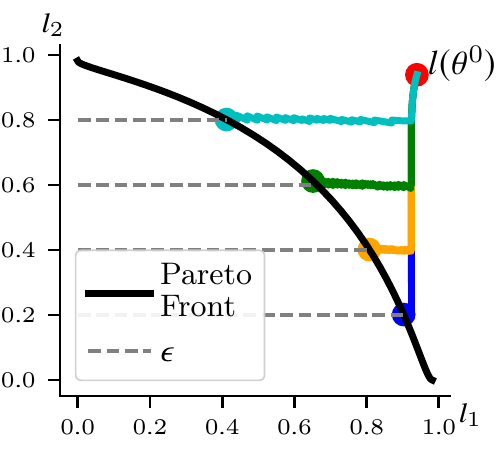}
    \label{fig:synours0}
  }%
  \subfigure[$\theta^{0}$ satisfies the constraints]{
    \centering
    \includegraphics[width=0.25\columnwidth]{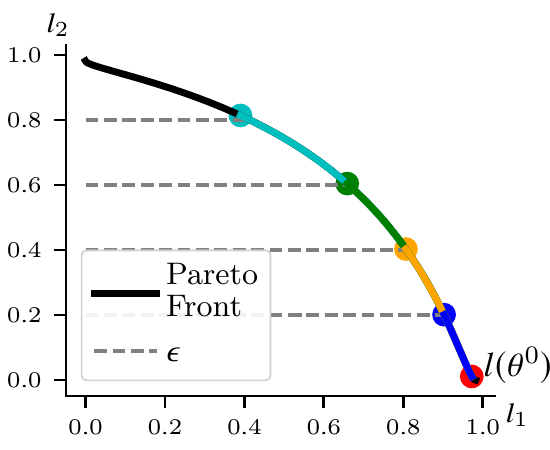}
    \label{fig:synours1}
  }%
  \setlength{\abovecaptionskip}{0pt}
  \setlength{\belowcaptionskip}{-10pt}
  \caption{Optimization trajectories of \method~in $n=20$ dimensional solution space ($\theta \in \mathbb{R}^{20}$). The initialization violates fairness constrains (left) and satisfies fairness constraints (right).}
  \label{fig:results1}
  \vspace{-.5cm}
\end{wrapfigure}

\subsection{Experiments on Synthetic Dataset}
Following the setting in \cite{lin2019pareto,mahapatra2020multi}, the synthetic data is from the two non-convex objectives to be minimized in Eq.(\ref{eq:syn1}) and the Pareto Front of the two objectives is also non-convex.
\begin{equation}
  \begin{aligned}
    & l_{1}(\theta)=1-e^{-\left\|\theta-\frac{1}{\sqrt{n}}\right\|_{2}^{2}} \\
    & l_{2}(\theta)=1-e^{-\left\|\theta+\frac{1}{\sqrt{n}}\right\|_{2}^{2}}.
  \end{aligned}
  \label{eq:syn1}
\end{equation}
Non-convex Pareto Front means that linear scalarization methods (e.g., FedAve) miss any solution in the concave part of the Pareto Front. In this experiment, we optimize $l_1$ under the constraint $l_2 \leq \epsilon, \epsilon \in \left\{0.2, 0.4, 0.6, 0.8 \right\}$. Considering the effect of the initialization in our experiment, we conduct experiments when the initialization $\theta^{0}$ satisfies the constraints and violates the constraints.

From the results in Figure~\ref{fig:results1}, when the initialization $\theta_{0}$ violates the constraints in Figure~\ref{fig:synours0}, the objective $l_2$ decreases in each step until it satisfies the constraint and finally \method~ reaches the constrained optimal $l_1(h^{*})$. As the initialization $\theta_{0}$ satisfies the constraints in Figure~\ref{fig:synours1}, our method focuses on optimizing $l_1$ until it achieves the optimal $l_1(h^{*})$ with the constraint $l_2 \leq \epsilon$.

\subsection{Experiments on Real-world Datasets with Equal Fairness Budgets}

\begin{figure}[h!]
  \setlength{\abovecaptionskip}{-0.1cm}
  \setlength{\belowcaptionskip}{-0.2cm}
  \centering
  \subfigure[]{
    \centering
    \includegraphics[width=0.23\columnwidth]{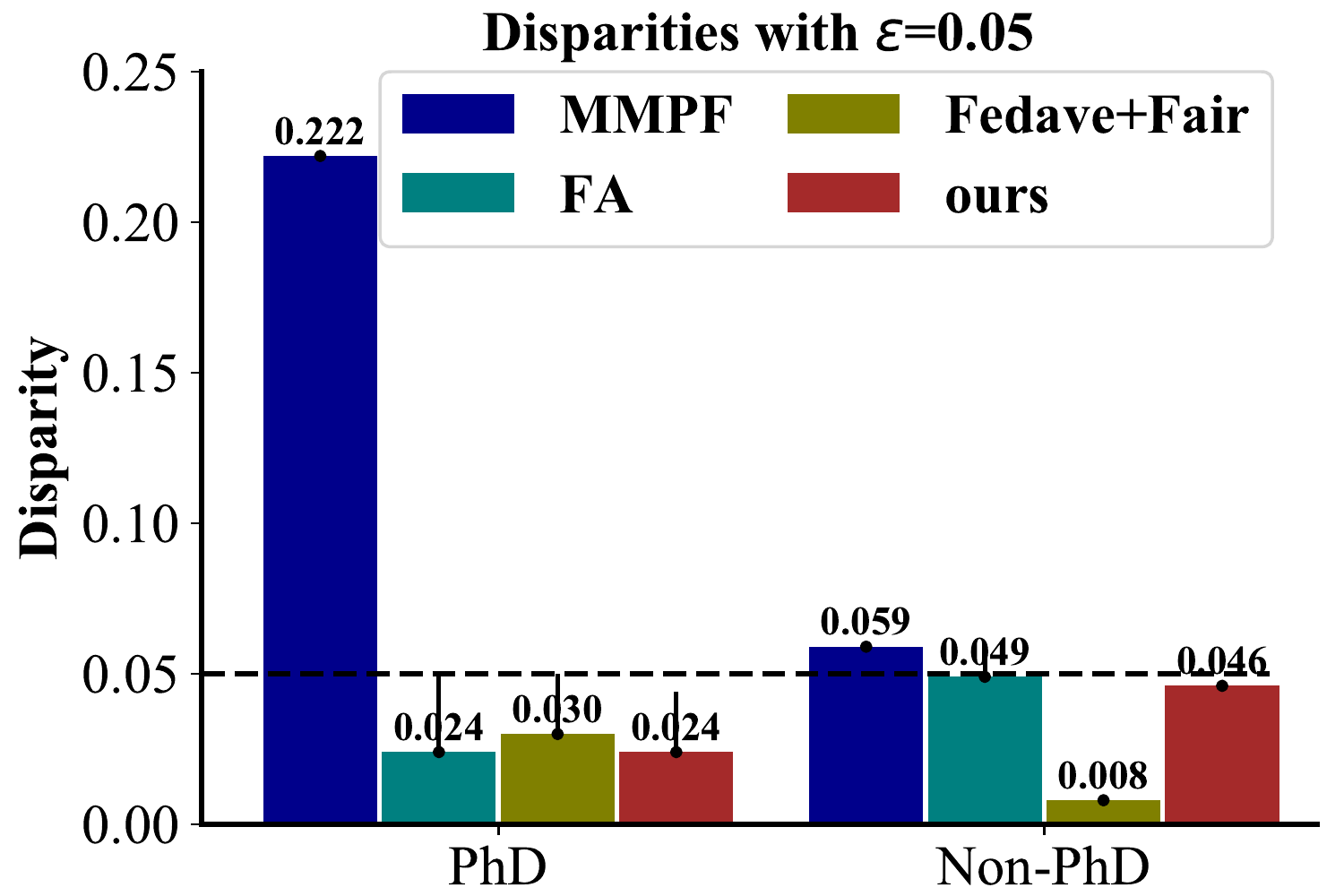}
  }%
  \subfigure[]{
    \centering
    \includegraphics[width=0.23\columnwidth]{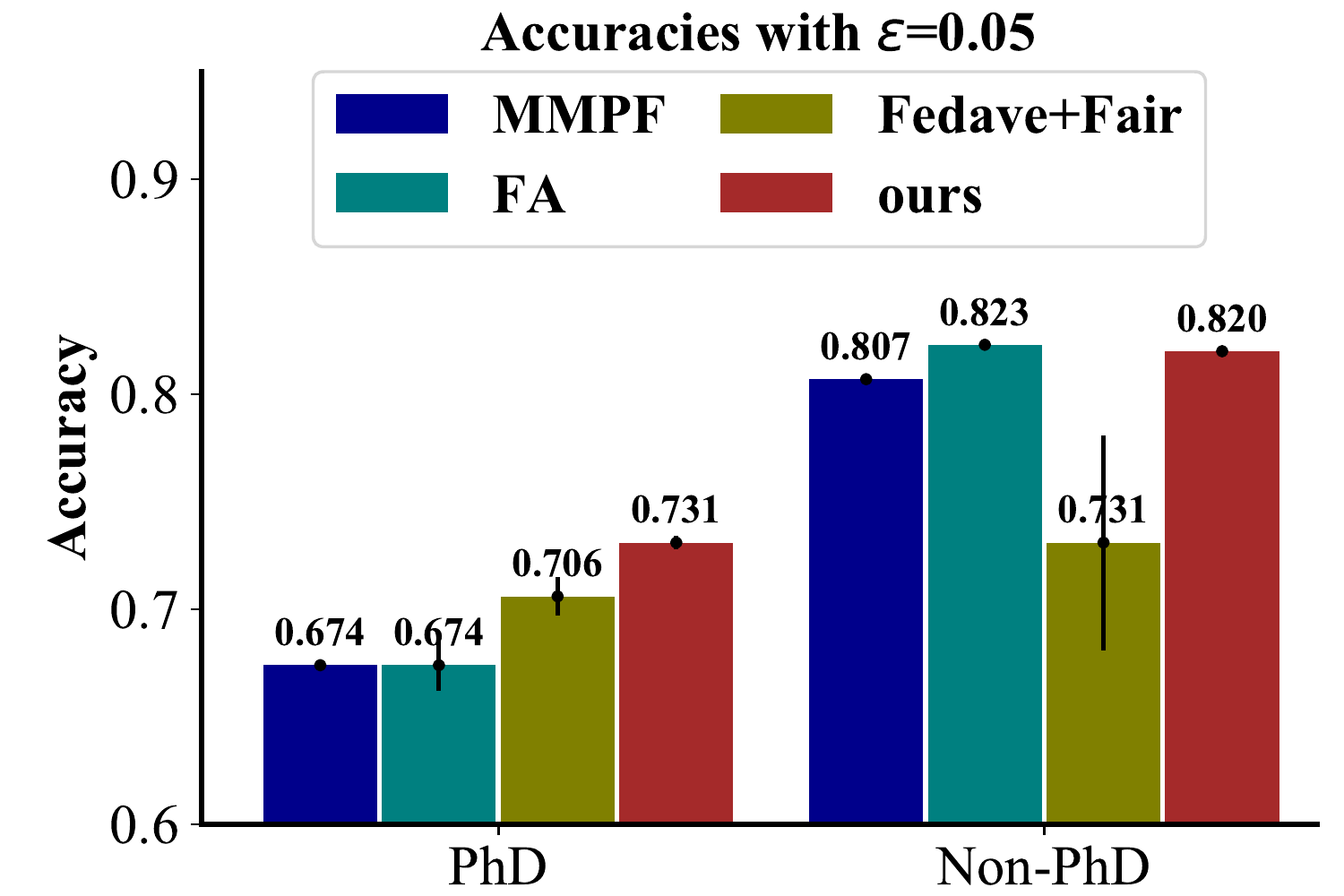}
  }%
  \subfigure[]{
    \centering
    \includegraphics[width=0.23\columnwidth]{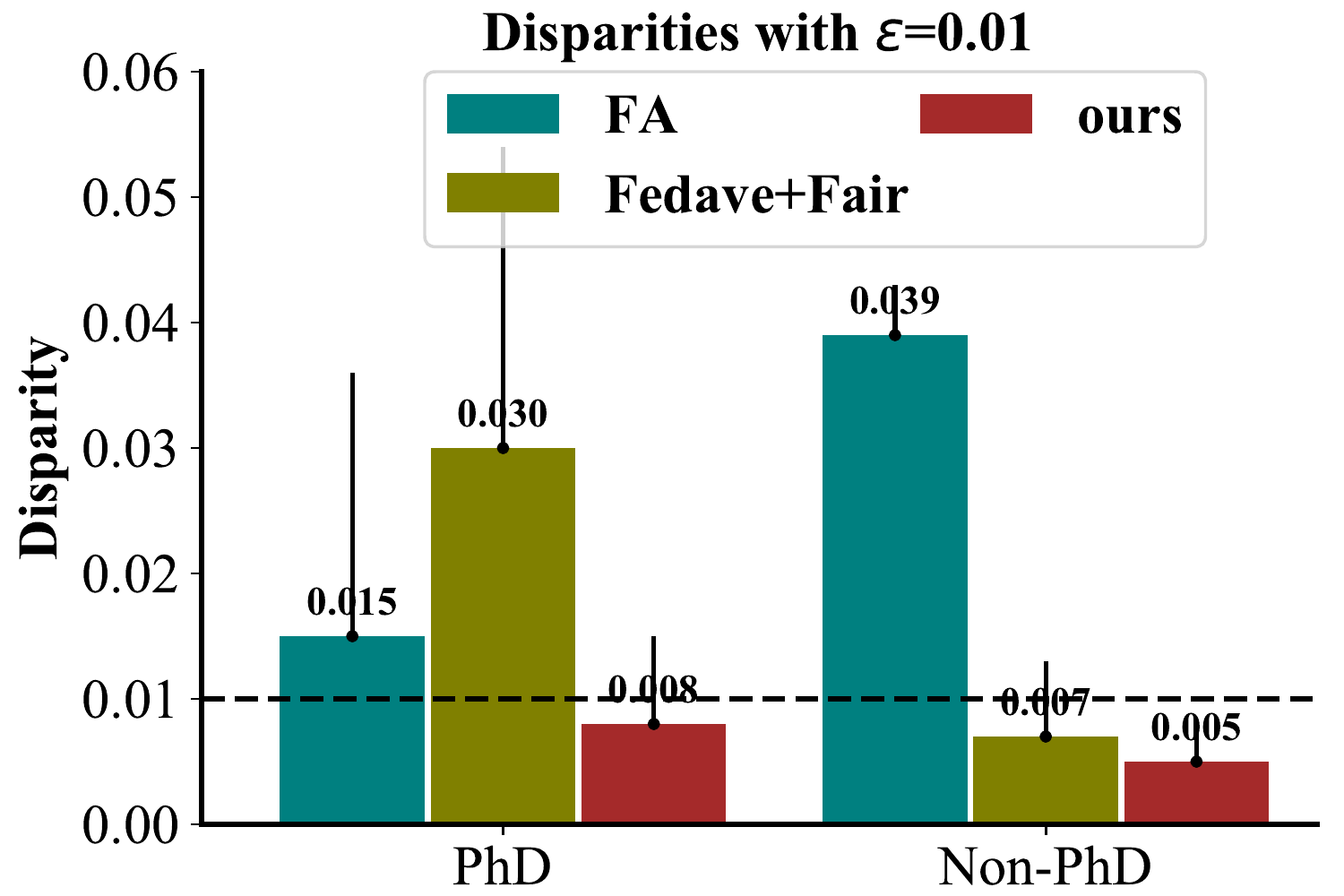}
  }%
  \subfigure[]{
    \centering
    \includegraphics[width=0.23\columnwidth]{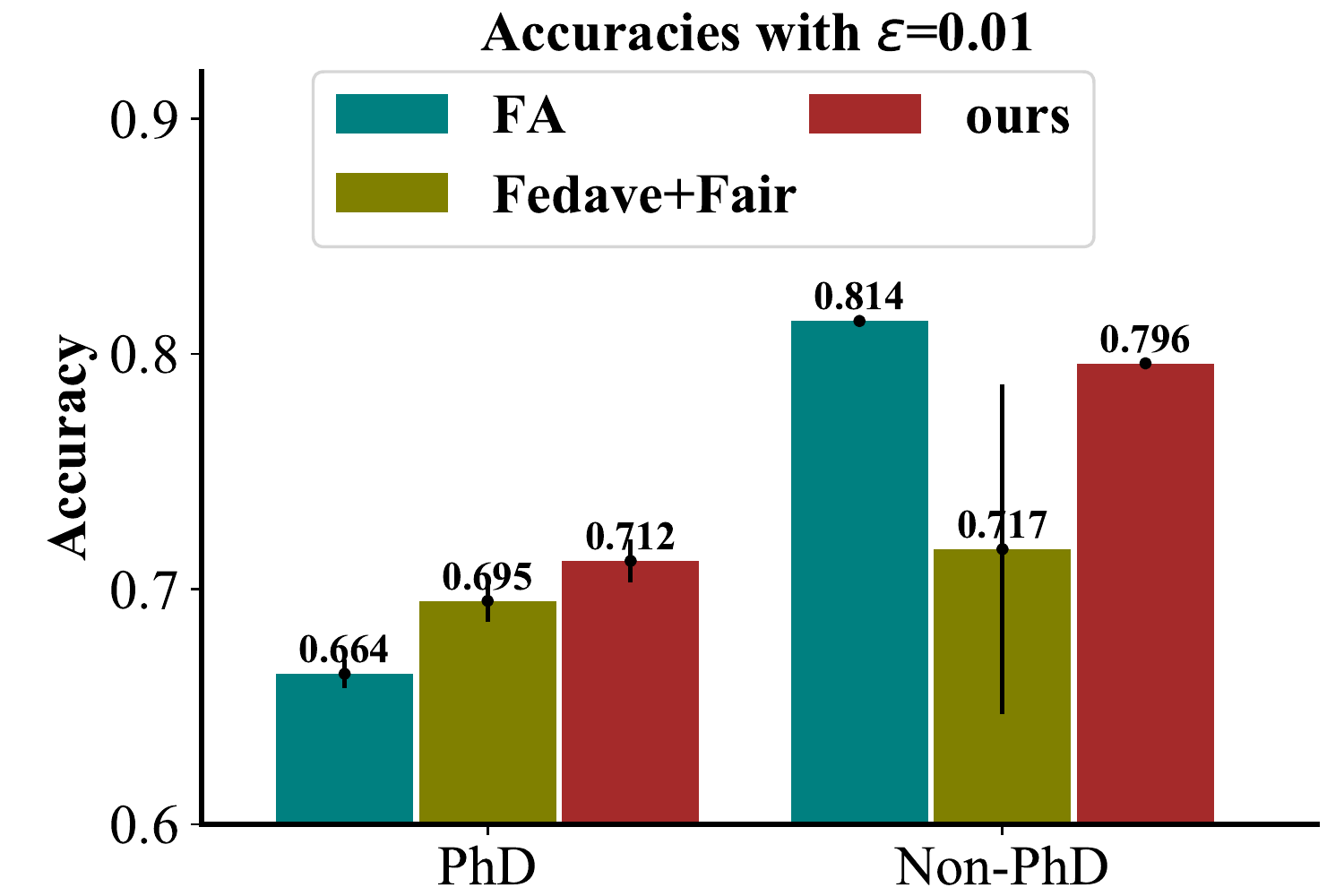}
  }%
  \caption{The disparities and accuracies on both clients as $\epsilon=0.05$ (top) and as $\epsilon=0.01$ of on Adult dataset when \emph{race} is the sensitive attribute.}
  \label{fig:uniform_adult_race}

\end{figure}

\subsubsection{Income Prediction on Adult Dataset}
We show the results with the sensitive attribute being \emph{race} in our main text and the results when \emph{gender} is the sensitive attribute are in Appendix. We set the fairness budgets defined in Eq.(\ref{eq:faircriteria_FL}) in two different cases, (1) looser constraint: $\Delta DP_{i} \leq \epsilon_{i} = 0.05 \ \forall i \in \left\{1,.,N\right\}$; (2) stricter constraint: $\Delta DP_{i} \leq \epsilon_{i} = 0.01 \ \forall i \in \left\{1,.,N\right\}$.


From Figure \ref{fig:uniform_adult_race}, \method~ achieves min-max performance on PhD client in the two cases with MCF. MMPF fails to achieve MCF as $\epsilon_{i} = 0.05$. FA and FedAve+FairReg achieve MCF with $\epsilon_{i} = 0.05$ but violate fairness constraint as $\epsilon_{i} = 0.01$. From Figure \ref{fig:uniform_adult_race}, our method achieves a comparable performance on non-PhD client compared to baselines.


\subsubsection{Prolonged Length of Stay Prediction on eICU Dataset}
The length of the patient's hospitalization is critical for the arrangement of the medical institutions as it is related to the allocation of limited ICU resources. We conduct experiments on the prediction of prolonged length of stay(whether the ICU stay is longer than 1 week) on eICU dataset. We use \emph{race} as sensitive attribute and set the fairness budgets defined in Eq.(\ref{eq:faircriteria_FL}) in two cases: (1) looser constraint: $ \Delta DP_{i} \leq \epsilon_{i} = 0.1 \ \forall i \in \left\{1,.,N\right\}$;  (2) stricter constraint: $\Delta DP_{i} \leq \epsilon_{i} = 0.05 \ \forall i \in \left\{1,.,N\right\}$

\begin{figure}[h!]
  \centering
  \subfigure[MMPF]{
    \centering
    \includegraphics[width=0.19\columnwidth]{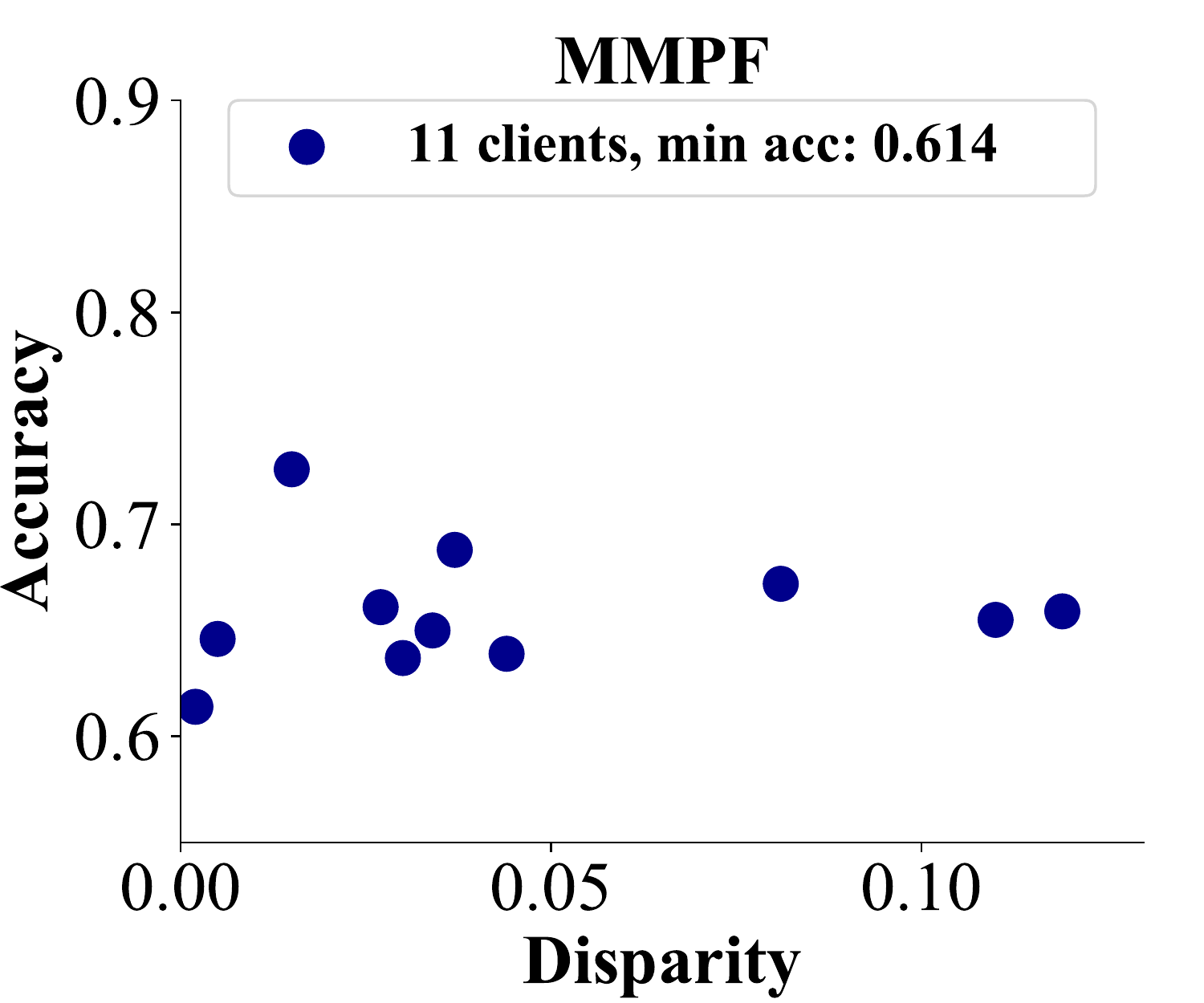}
  }%
  \subfigure[FA]{
    \centering
    \includegraphics[width=0.19\columnwidth]{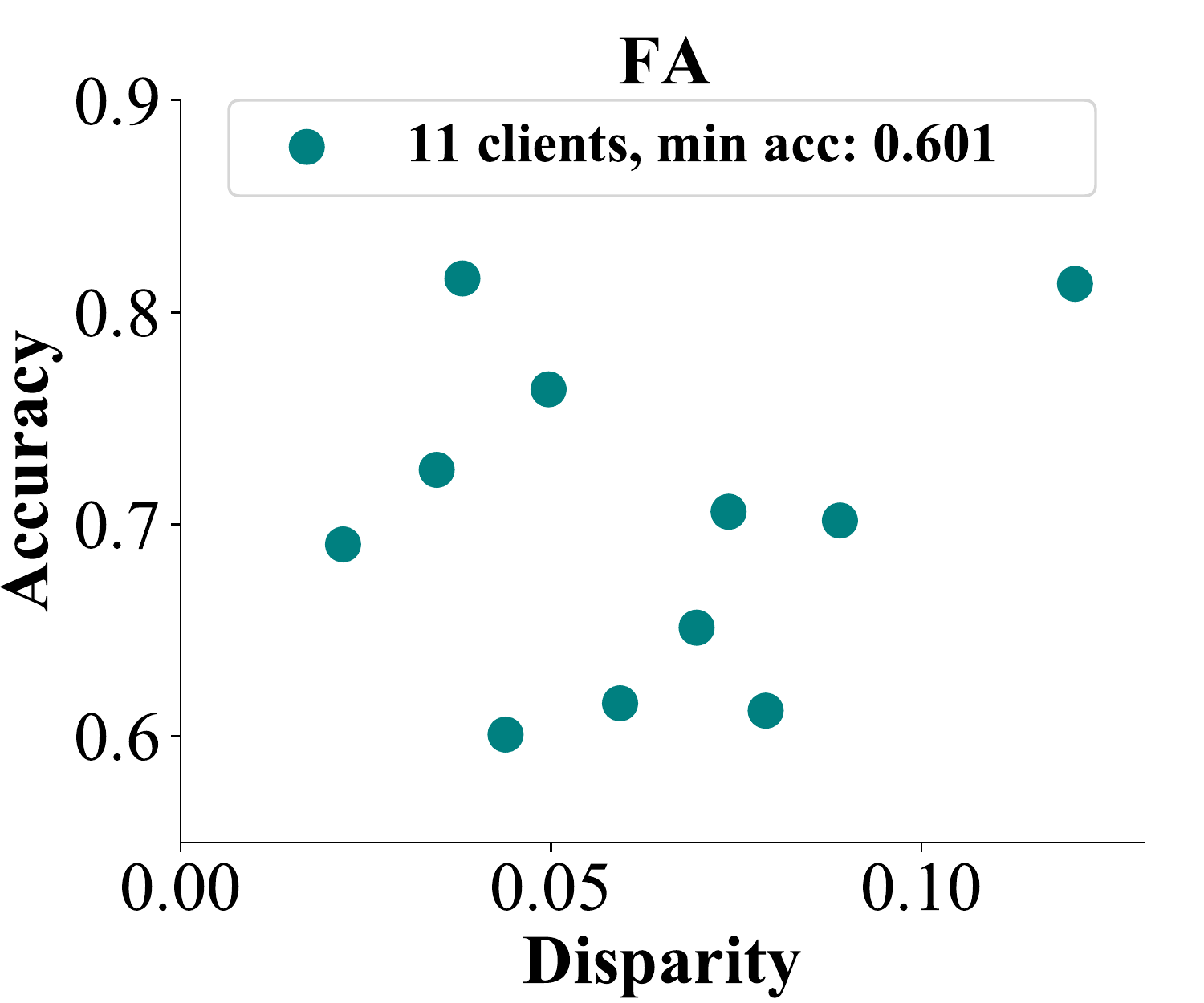}
  }%
  \subfigure[FedAve+FairReg]{
    \centering
    \includegraphics[width=0.19\columnwidth]{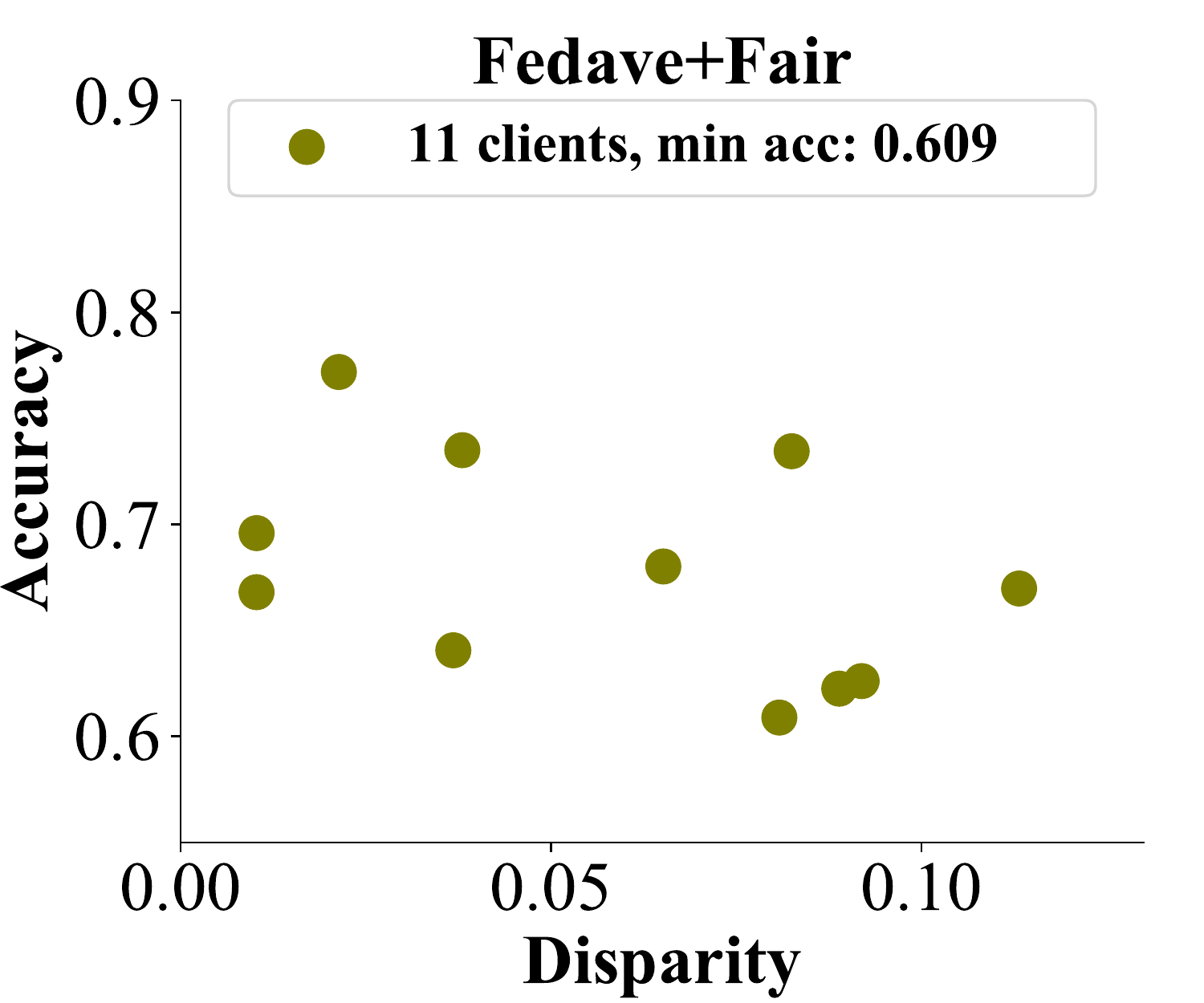}
  }%
  \subfigure[ours with $\epsilon=0.1$]{
    \centering
    \includegraphics[width=0.19\columnwidth]{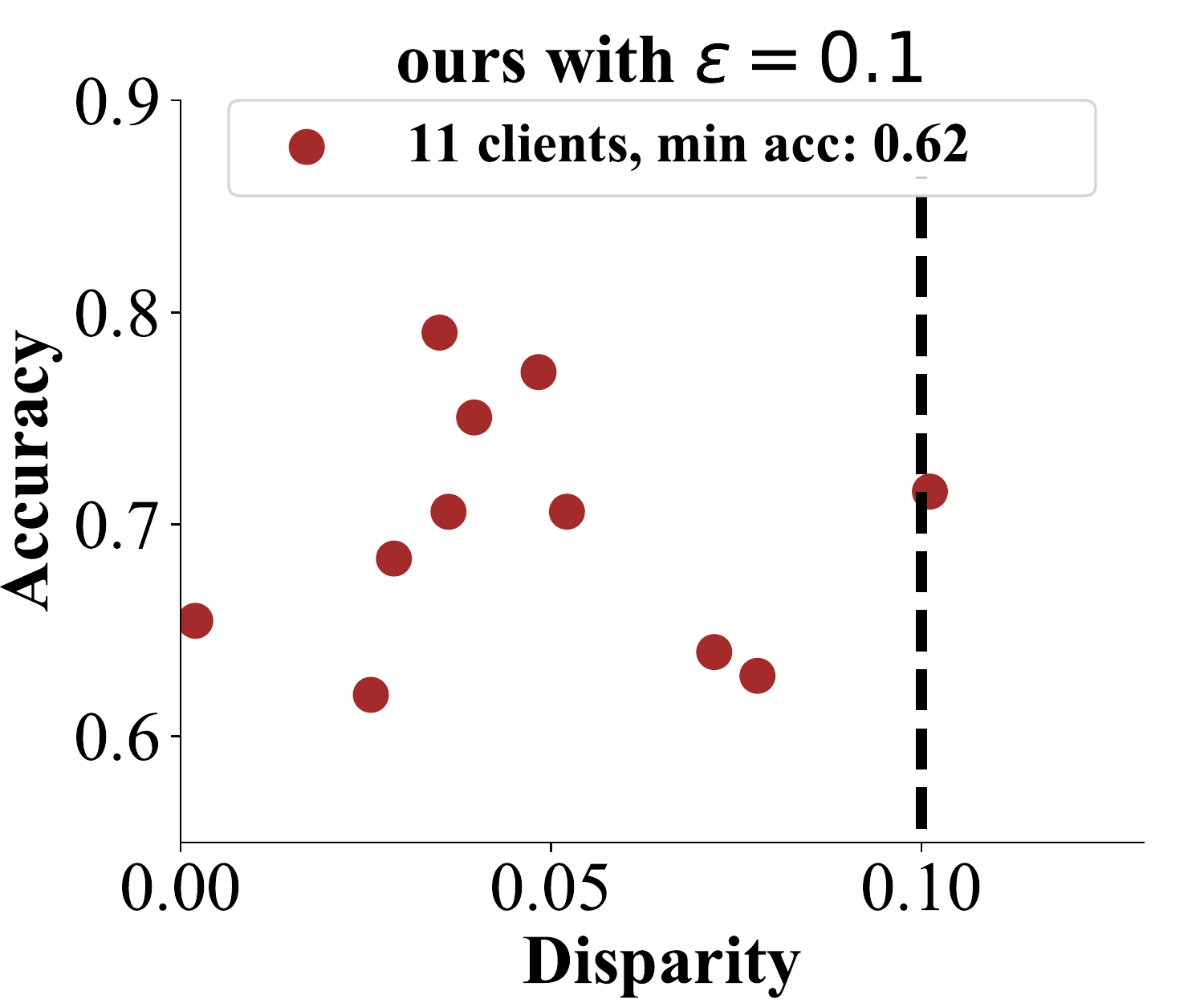}
  }%
  \subfigure[ours with $\epsilon=0.05$]{
    \centering
    \includegraphics[width=0.19\columnwidth]{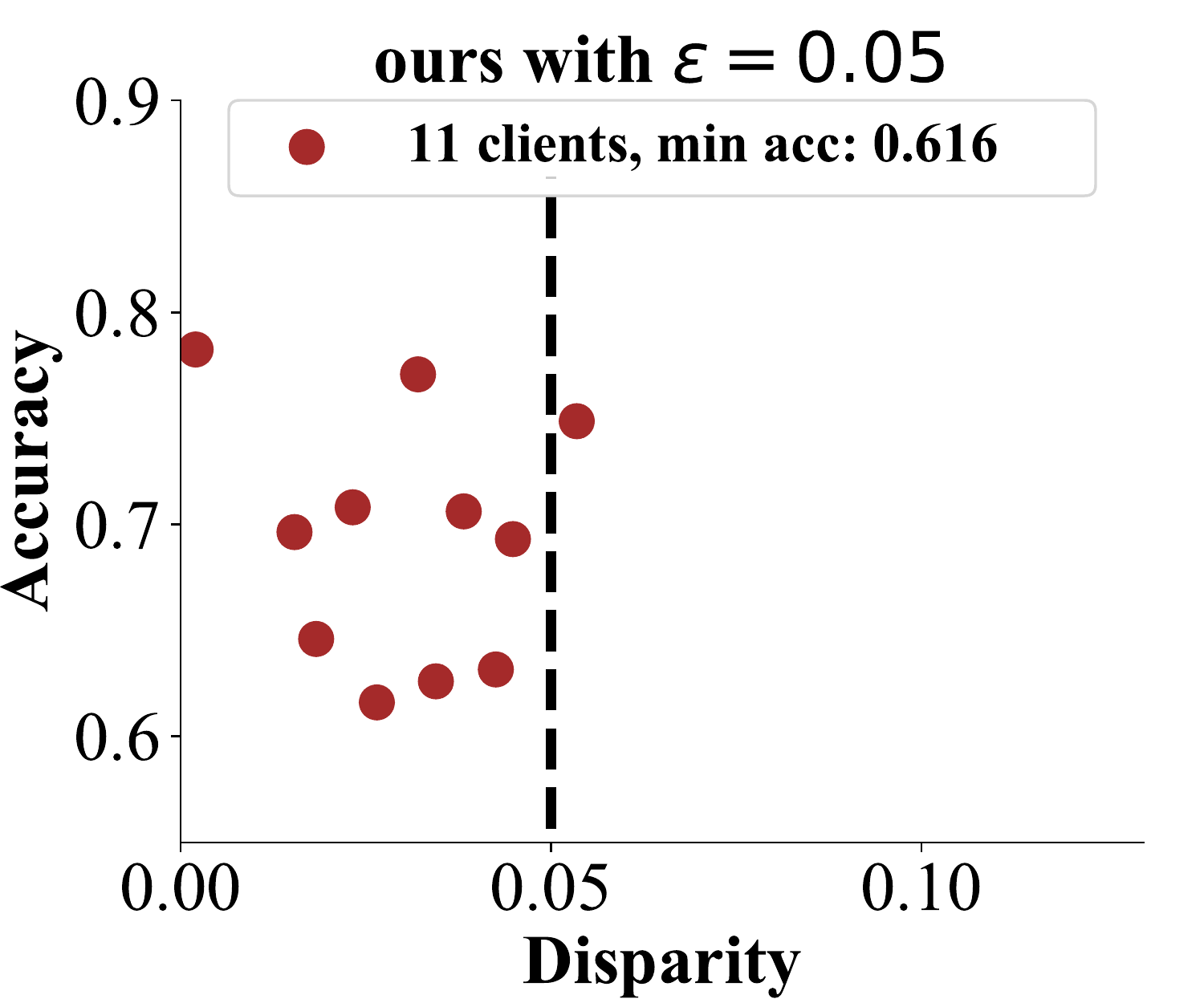}
  }%
  \setlength{\abovecaptionskip}{0pt}
  \setlength{\belowcaptionskip}{-0pt}
  \caption{Experiments on LoS Prediction task with the sensitive attribute being \emph{race}. The points in each figure denote the clients and X and Y coordinates of the points denote the disparities and the accuracies, respectively.}
  \label{fig:uniform_eicu_race}
\end{figure}

From Figure \ref{fig:uniform_eicu_race}, our method achieves min-max performance with fairness budget $\epsilon = 0.1$ compared to the baselines. When we constrain $\Delta DP_{i} \leq 0.05 \ \forall i$, all baselines fail to satisfy the constraints and the disparities are about 0.1 while our method significantly decreases the disparities and the maximum of the disparities is $0.68$. Besides, we maintain comparable utilities on all clients compared to baselines.

\subsection{Experiments with Client-Specific Fairness Budgets}
Data heterogeneity encourages the different levels of the disparities on different clients. Consistent fairness budgets can cause unexpected hurt on some specific clients with severe disparities. We explore the performance of our method given client-specific fairness budgets. Specifically, we firstly conduct unconstrained min-max experiments and measure the original disparities $\Delta \bm{DP} = [\Delta DP_{1},.,\Delta DP_{N}]$ on all clients, then we constrain the model disparities based on the original disparities of all clients, i.e., $[\epsilon_{1}, \epsilon_{2}, ., \epsilon_{N}] = w \cdot \Delta \bm{DP}, w \in \left\{1.0, 0.9, 0.8, ., 0.2\right\}$.

\begin{figure}[h!]
  \vspace{-.2cm}
  \centering
  \subfigure[\emph{Adult} $Disparity-w$]{
    \centering
    \includegraphics[width=0.23\columnwidth]{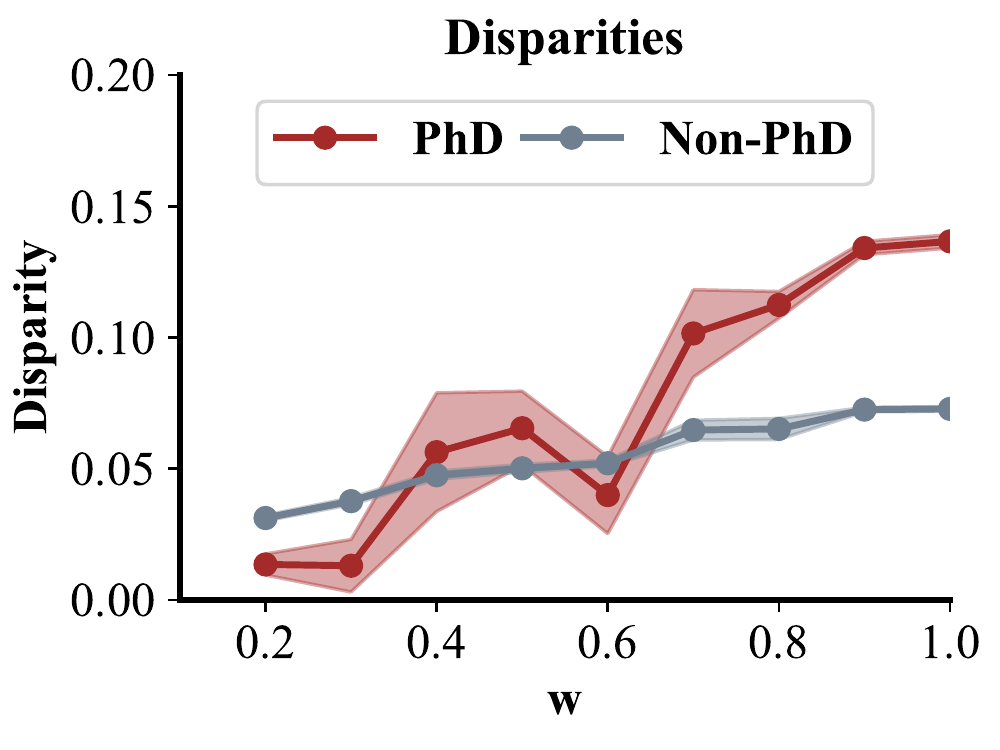}
    \label{fig:specific_adult_race_dis}
  }%
  \subfigure[\emph{Adult} $Accuracy-w$]{
    \centering
    \includegraphics[width=0.23\columnwidth]{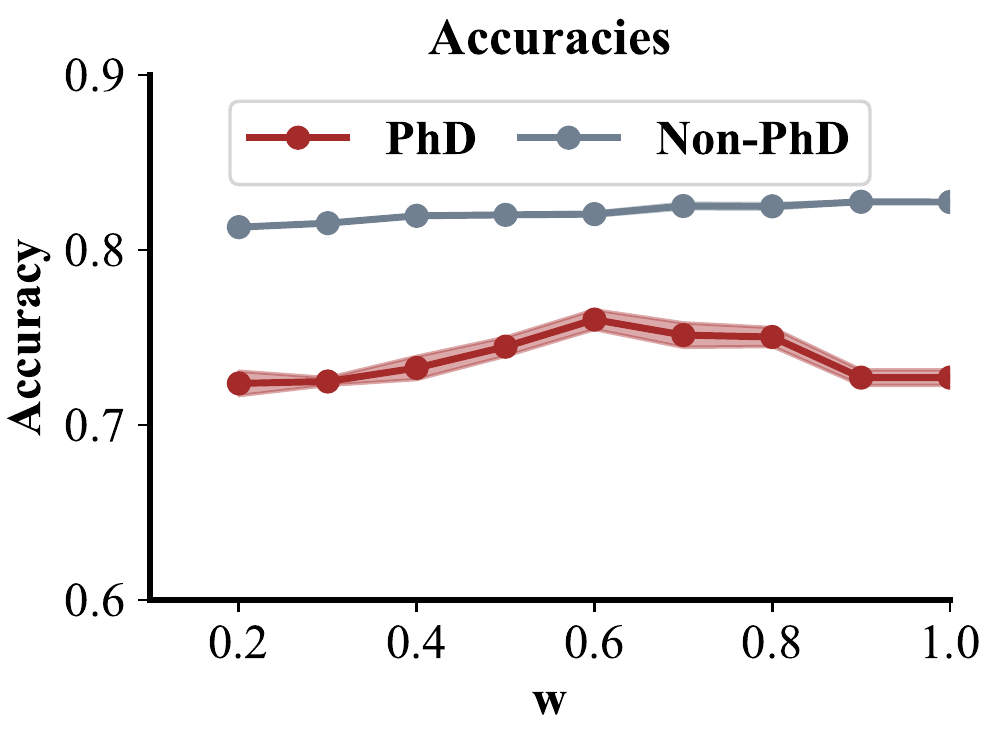}
    \label{fig:specific_adult_race_acc}
  }%
  \subfigure[\emph{eICU} $Disparity-w$]{
    \centering
    \includegraphics[width=0.23\columnwidth]{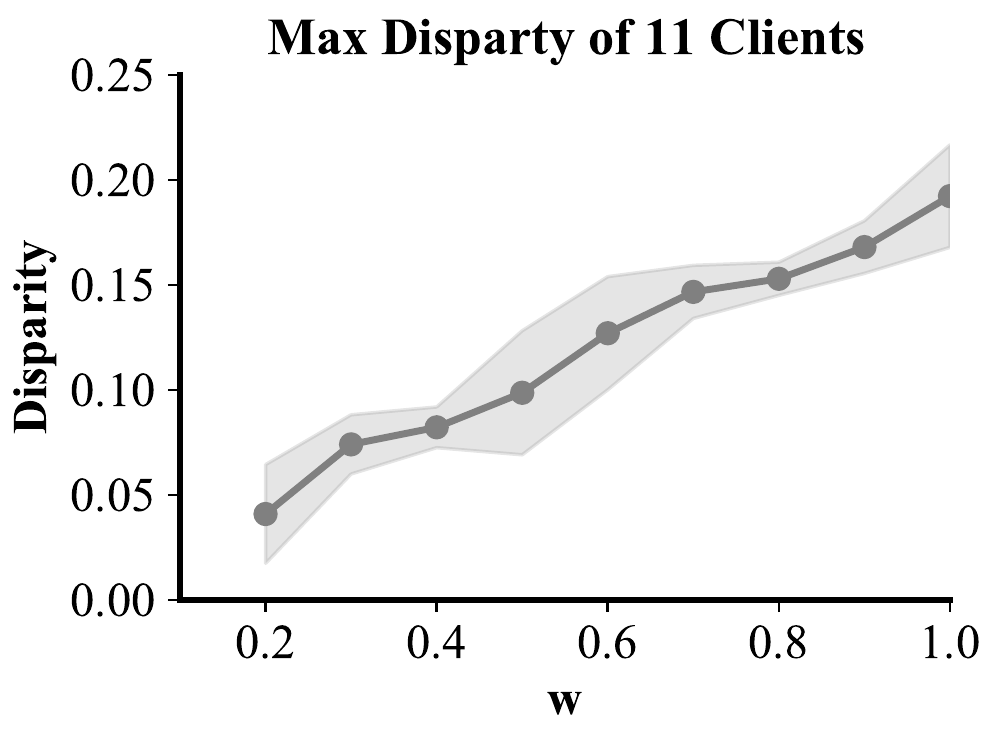}
    \label{fig:specific_eicu_dis}
  }%
  \subfigure[\emph{eICU} $Accuracy-w$]{
    \centering
    \includegraphics[width=0.23\columnwidth]{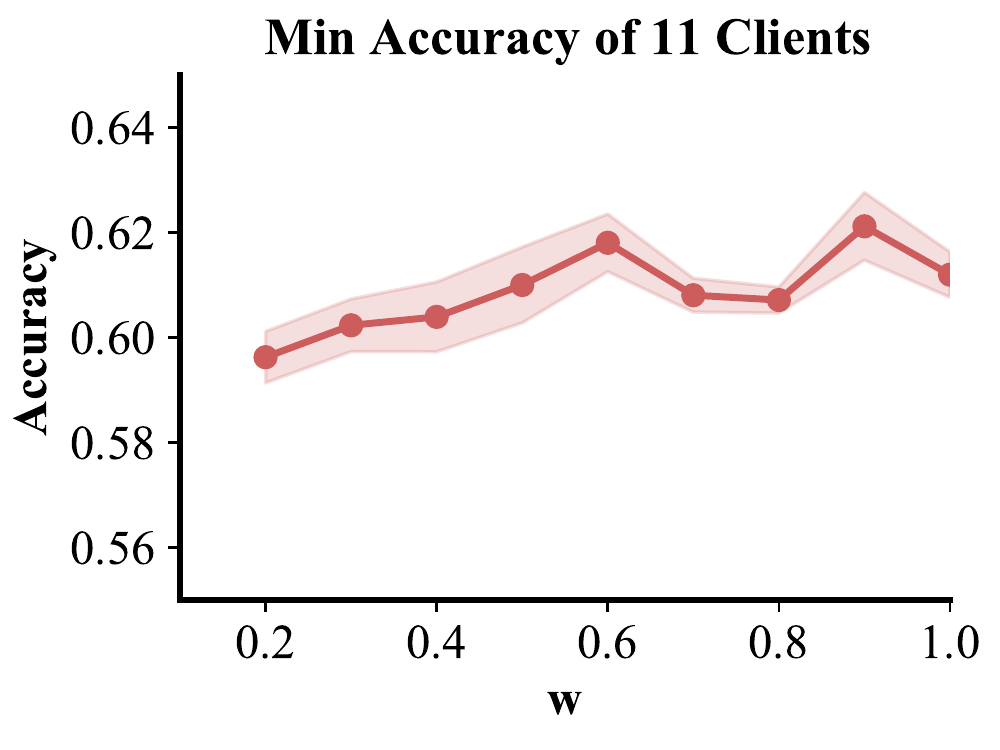}
    \label{fig:specific_eicu_acc}
  }%
  \caption{Client-specific constraint experiment on Adult and eICU.}
  \label{fig:specific_adult_race_result}
  \vspace{-.2cm}
\end{figure}



From Figure~\ref{fig:specific_adult_race_dis} and Figure~\ref{fig:specific_adult_race_acc}, we show the effect of the client-specific fairness budgets on model disparities and utilities. The decreasing $w$ means the stricter constraints on both clients and \method~ reduces the disparities significantly as shown in Figure~\ref{fig:specific_adult_race_dis}. With the stricter client-specific constraints on both clients, the utilities on both clients decrease slightly as shown in Figure~\ref{fig:specific_adult_race_acc}, which implies that \method~achieves a great balance between the fairness and the utility of all clients. \method~is compatible with client-specific fairness budgets which enhances its flexibility and avoids severe hurt to the specific clients.


The results of LoS prediction task with client-specific fairness budgets are shown in Figure \ref{fig:specific_eicu_dis} and Figure~\ref{fig:specific_eicu_acc}. As $w$ decreases from 1.0 to 0.2, the maximum of all client disparities in Figure \ref{fig:specific_eicu_dis} decrease from 0.2 to 0.05 which means the model becomes fairer on all clients. Figure \ref{fig:specific_eicu_acc} shows the minimum of the utilities which slightly decreases from 0.62 to 0.6 and our method achieves an acceptable trade-off between model fairness and utility as the amount of clients increases in this task.

\section{Conclusion}
In this paper, we investigate the consistency and fairness issues in federated networks as the learned model deployed on local clients can cause inconsistent performances and disparities without elaborate design. We propose a novel method called \method~to overcome the disparity and inconsistency concerns in the favored direction of gradient-based constrained multi-objective optimization. Comprehensive empirical evaluation results measured by quantitative metrics demonstrate the effectiveness, superiority, and reliability of our proposed method.

\section*{Acknowledgments}
Sen Cui, Weishen Pan and Changshui Zhang would like to acknowledge the funding by the National Key Research and Development Program of China (No. 2018AAA0100701). Fei Wang acknowledges the support from AWS Machine Learning for Research Award and Google Faculty Research Award.

\small
\bibliographystyle{unsrt}
\bibliography{neurips_refs}

\begin{thebibliography}{10}

\bibitem{yang2019federated}
Qiang Yang, Yang Liu, Yong Cheng, Yan Kang, Tianjian Chen, and Han Yu.
\newblock Federated learning.
\newblock {\em Synthesis Lectures on Artificial Intelligence and Machine
  Learning}, 13(3):1--207, 2019.

\bibitem{xu2020federated}
Jie Xu, Benjamin~S Glicksberg, Chang Su, Peter Walker, Jiang Bian, and Fei
  Wang.
\newblock Federated learning for healthcare informatics.
\newblock {\em Journal of Healthcare Informatics Research}, pages 1--19, 2020.

\bibitem{angwin2016machine}
Julia Angwin, Jeff Larson, Surya Mattu, and Lauren Kirchner.
\newblock Machine bias: There’s software used across the country to predict
  future criminals.
\newblock {\em And it’s biased against blacks. ProPublica}, 23, 2016.

\bibitem{li2019fair}
Tian Li, Maziar Sanjabi, Ahmad Beirami, and Virginia Smith.
\newblock Fair resource allocation in federated learning.
\newblock In {\em International Conference on Learning Representations}, 2019.

\bibitem{mohri2019agnostic}
Mehryar Mohri, Gary Sivek, and Ananda~Theertha Suresh.
\newblock Agnostic federated learning.
\newblock In {\em International Conference on Machine Learning}, pages
  4615--4625, 2019.

\bibitem{du2020fairness}
Wei Du, Depeng Xu, Xintao Wu, and Hanghang Tong.
\newblock Fairness-aware agnostic federated learning.
\newblock {\em arXiv preprint arXiv:2010.05057}, 2020.

\bibitem{vaid2021federated}
Akhil Vaid, Suraj~K Jaladanki, Jie Xu, Shelly Teng, Arvind Kumar, Samuel Lee,
  Sulaiman Somani, Ishan Paranjpe, Jessica~K De~Freitas, Tingyi Wanyan, et~al.
\newblock Federated learning of electronic health records improves mortality
  prediction in patients hospitalized with covid-19.
\newblock {\em JMIR medical informatics}, 2021.

\bibitem{dwork2012fairness}
Cynthia Dwork, Moritz Hardt, Toniann Pitassi, Omer Reingold, and Richard Zemel.
\newblock Fairness through awareness.
\newblock In {\em Proceedings of the 3rd innovations in theoretical computer
  science conference}, pages 214--226, 2012.

\bibitem{hardt2016equality}
Moritz Hardt, Eric Price, and Nathan Srebro.
\newblock Equality of opportunity in supervised learning.
\newblock In {\em Proceedings of the 30th International Conference on Neural
  Information Processing Systems}, pages 3323--3331, 2016.

\bibitem{mcmahan2017communication}
Brendan McMahan, Eider Moore, Daniel Ramage, Seth Hampson, and Blaise~Aguera
  y~Arcas.
\newblock Communication-efficient learning of deep networks from decentralized
  data.
\newblock In {\em Artificial Intelligence and Statistics}, pages 1273--1282.
  PMLR, 2017.

\bibitem{pan2021explaining}
Weishen Pan, Sen Cui, Jiang Bian, Changshui Zhang, and Fei Wang.
\newblock Explaining algorithmic fairness through fairness-aware causal path
  decomposition.
\newblock In {\em Proceedings of the 27th ACM SIGKDD Conference on Knowledge
  Discovery \& Data Mining}, pages 1287--1297, 2021.

\bibitem{cui2021towards}
Sen Cui, Weishen Pan, Changshui Zhang, and Fei Wang.
\newblock Towards model-agnostic post-hoc adjustment for balancing ranking
  fairness and algorithm utility.
\newblock In {\em Proceedings of the 27th ACM SIGKDD Conference on Knowledge
  Discovery \& Data Mining}, pages 207--217, 2021.

\bibitem{kallus2019fairness}
Nathan Kallus and Angela Zhou.
\newblock The fairness of risk scores beyond classification: Bipartite ranking
  and the xauc metric.
\newblock In {\em Advances in Neural Information Processing Systems}, pages
  3433--3443, 2019.

\bibitem{beutel2017data}
Alex Beutel, Jilin Chen, Zhe Zhao, and Ed~H Chi.
\newblock Data decisions and theoretical implications when adversarially
  learning fair representations.
\newblock {\em arXiv preprint arXiv:1707.00075}, 2017.

\bibitem{louizos2015variational}
Christos Louizos, Kevin Swersky, Yujia Li, Max Welling, and Richard Zemel.
\newblock The variational fair autoencoder.
\newblock {\em arXiv preprint arXiv:1511.00830}, 2015.

\bibitem{madras2018learning}
David Madras, Elliot Creager, Toniann Pitassi, and Richard Zemel.
\newblock Learning adversarially fair and transferable representations.
\newblock In {\em International Conference on Machine Learning}, pages
  3384--3393, 2018.

\bibitem{zemel2013learning}
Rich Zemel, Yu~Wu, Kevin Swersky, Toni Pitassi, and Cynthia Dwork.
\newblock Learning fair representations.
\newblock In {\em International Conference on Machine Learning}, pages
  325--333, 2013.

\bibitem{zhang2018mitigating}
Brian~Hu Zhang, Blake Lemoine, and Margaret Mitchell.
\newblock Mitigating unwanted biases with adversarial learning.
\newblock In {\em Proceedings of the 2018 AAAI/ACM Conference on AI, Ethics,
  and Society}, pages 335--340, 2018.

\bibitem{kamishima2011fairness}
Toshihiro Kamishima, Shotaro Akaho, and Jun Sakuma.
\newblock Fairness-aware learning through regularization approach.
\newblock In {\em 2011 IEEE 11th International Conference on Data Mining
  Workshops}, pages 643--650. IEEE, 2011.

\bibitem{zafar2015fairness}
Muhammad~Bilal Zafar, Isabel Valera, Manuel~Gomez Rodriguez, and Krishna~P
  Gummadi.
\newblock Fairness constraints: Mechanisms for fair classification.
\newblock {\em arXiv preprint arXiv:1507.05259}, 2015.

\bibitem{beutel2019putting}
Alex Beutel, Jilin Chen, Tulsee Doshi, Hai Qian, Allison Woodruff, Christine
  Luu, Pierre Kreitmann, Jonathan Bischof, and Ed~H Chi.
\newblock Putting fairness principles into practice: Challenges, metrics, and
  improvements.
\newblock In {\em Proceedings of the 2019 AAAI/ACM Conference on AI, Ethics,
  and Society}, pages 453--459, 2019.

\bibitem{chen2018gradnorm}
Zhao Chen, Vijay Badrinarayanan, Chen-Yu Lee, and Andrew Rabinovich.
\newblock Gradnorm: Gradient normalization for adaptive loss balancing in deep
  multitask networks.
\newblock In {\em International Conference on Machine Learning}, pages
  794--803. PMLR, 2018.

\bibitem{mahapatra2020multi}
Debabrata Mahapatra and Vaibhav Rajan.
\newblock Multi-task learning with user preferences: Gradient descent with
  controlled ascent in pareto optimization.
\newblock In {\em International Conference on Machine Learning}, pages
  6597--6607. PMLR, 2020.

\bibitem{martinez2020minimax}
Natalia Martinez, Martin Bertran, and Guillermo Sapiro.
\newblock Minimax pareto fairness: A multi objective perspective.
\newblock In {\em International Conference on Machine Learning}, pages
  6755--6764. PMLR, 2020.

\bibitem{coello2006evolutionary}
CA~Coello Coello.
\newblock Evolutionary multi-objective optimization: a historical view of the
  field.
\newblock {\em IEEE computational intelligence magazine}, 1(1):28--36, 2006.

\bibitem{evtushenko2014deterministic}
Yu~G Evtushenko and MA~Posypkin.
\newblock A deterministic algorithm for global multi-objective optimization.
\newblock {\em Optimization Methods and Software}, 29(5):1005--1019, 2014.

\bibitem{zerbinati2011comparison}
Adrien Zerbinati, Jean-Antoine Desideri, and R{\'e}gis Duvigneau.
\newblock Comparison between mgda and paes for multi-objective optimization.
\newblock 2011.

\bibitem{desideri2012multiple}
Jean-Antoine D{\'e}sid{\'e}ri.
\newblock Multiple-gradient descent algorithm (mgda) for multiobjective
  optimization.
\newblock {\em Comptes Rendus Mathematique}, 350(5-6):313--318, 2012.

\bibitem{polak2003algorithms}
E~Polak, JO~Royset, and RS~Womersley.
\newblock Algorithms with adaptive smoothing for finite minimax problems.
\newblock {\em Journal of Optimization Theory and Applications},
  119(3):459--484, 2003.

\bibitem{lin2019pareto}
Xi~Lin, Hui-Ling Zhen, Zhenhua Li, Qing-Fu Zhang, and Sam Kwong.
\newblock Pareto multi-task learning.
\newblock In {\em Advances in Neural Information Processing Systems}, pages
  12060--12070, 2019.

\bibitem{pollard2018eicu}
Tom~J Pollard, Alistair~EW Johnson, Jesse~D Raffa, Leo~A Celi, Roger~G Mark,
  and Omar Badawi.
\newblock The eicu collaborative research database, a freely available
  multi-center database for critical care research.
\newblock {\em Scientific data}, 5(1):1--13, 2018.

\bibitem{johnson2018generalizability}
Alistair E.~W. Johnson, Tom~J. Pollard, and Tristan Naumann.
\newblock Generalizability of predictive models for intensive care unit
  patients.
\newblock {\em Machine Learning for Health (ML4H) Workshop at NeurIPS 2018},
  2018.

\end{thebibliography}

\section*{Checklist}

The checklist follows the references.  Please
read the checklist guidelines carefully for information on how to answer these
questions.  For each question, change the default \answerTODO{} to \answerYes{},
\answerNo{}, or \answerNA{}.  You are strongly encouraged to include a {\bf
justification to your answer}, either by referencing the appropriate section of
your paper or providing a brief inline description.  For example:
\begin{itemize}
  \item Did you include the license to the code and datasets? \answerYes{}
  \item Did you include the license to the code and datasets? \answerNo{The code and the data are proprietary.}
  \item Did you include the license to the code and datasets? \answerNA{}
\end{itemize}
Please do not modify the questions and only use the provided macros for your
answers.  Note that the Checklist section does not count towards the page
limit.  In your paper, please delete this instructions block and only keep the
Checklist section heading above along with the questions/answers below.

\begin{enumerate}

\item For all authors...
\begin{enumerate}
  \item Do the main claims made in the abstract and introduction accurately reflect the paper's contributions and scope?
    \answerYes{}
  \item Did you describe the limitations of your work?
    \answerYes{We describe the limitations in Appendix}
  \item Did you discuss any potential negative societal impacts of your work?
    \answerNA{We discuss it in Appendix}
  \item Have you read the ethics review guidelines and ensured that your paper conforms to them?
    \answerYes{}
\end{enumerate}

\item If you are including theoretical results...
\begin{enumerate}
  \item Did you state the full set of assumptions of all theoretical results?
    \answerYes{We state all assumptions in Appendix}
  \item Did you include complete proofs of all theoretical results?
    \answerYes{We include the complete proofs in Appendix}
\end{enumerate}

\item If you ran experiments...
\begin{enumerate}
  \item Did you include the code, data, and instructions needed to reproduce the main experimental results (either in the supplemental material or as a URL)?
    \answerYes{We made our source code publicly on github}
  \item Did you specify all the training details (e.g., data splits, hyperparameters, how they were chosen)?
    \answerYes{We discuss the training details in Appendix}
  \item Did you report error bars (e.g., with respect to the random seed after running experiments multiple times)?
    \answerYes{We report the error bars of all experimental results}
  \item Did you include the total amount of compute and the type of resources used (e.g., type of GPUs, internal cluster, or cloud provider)?
    \answerYes{We report the resources in Appendix}
\end{enumerate}

\item If you are using existing assets (e.g., code, data, models) or curating/releasing new assets...
\begin{enumerate}
  \item If your work uses existing assets, did you cite the creators?
    \answerYes{We cite the creators and discuss it in Appendix}
  \item Did you mention the license of the assets?
    \answerYes{We mention it in Appendix}
  \item Did you include any new assets either in the supplemental material or as a URL?
    \answerYes{We made our code publicly on github}
  \item Did you discuss whether and how consent was obtained from people whose data you're using/curating?
    \answerYes{We discuss it in Appendix}
  \item Did you discuss whether the data you are using/curating contains personally identifiable information or offensive content?
    \answerNA{}
\end{enumerate}

\item If you used crowdsourcing or conducted research with human subjects...
\begin{enumerate}
  \item Did you include the full text of instructions given to participants and screenshots, if applicable?
    \answerNA{}
  \item Did you describe any potential participant risks, with links to Institutional Review Board (IRB) approvals, if applicable?
    \answerNA{}
  \item Did you include the estimated hourly wage paid to participants and the total amount spent on participant compensation?
    \answerNA{}
\end{enumerate}

\newpage


\section*{Supplementary material}

# A  Algorithm Details

We summarize our method in Algorithm 1 as follows:

---
**Algorithm 1:** FCFL : obtain a **fair Pareto min-max** model
---
**Input:** fairness budget $\{\epsilon_i\}$, gradient threshold $\epsilon_d$, the initial parameters $\delta_l^0, \delta_g^0$, the learning rate $\eta$, the parameter decay rate $\beta$, the initial hypothesis $h_{\theta^0}$, and the data of all clients $\{D^1, D^2, ...D^N\}$.
**Stage 1:** obtain a fair min-max model $h^1$
**for** $t = 1, 2, ...T^1$ **do**
> *Local clients:*
> The center server sends the global model $h^t$ to all local client;
> Local clients evaluate the performance of $h^t$ and obtain the local gradients $\nabla_{\theta_t} l_i$;
> *Center server:*
> Calculate the surrogate maximum objective $\hat{l}$ and disparity $\hat{g}'$ using the Eq.( 10) in the main text
> **if** $\hat{g}' \leq 0$ **then**
> > Determine the gradient direction $d^t$ by solving the LP problem in the Eq.(12) in the main text;
>
> **end**
> **else**
> > Determine $d^t$ by solving the problem in the Eq.(13) in the main text;
>
> **end**
> Update the model $\theta^{t+1} = \theta^t + \eta d^t$;
> **if** $||d^t|| \leq \epsilon_d$ **then**
> > Decay the parameter $\delta_l^t$ and $\delta_g^t$: $\delta_l^{t+1} = \beta \cdot \delta_l^t$, $\delta_g^{t+1} = \beta \cdot \delta_g^t$
>
> **end**
> **else**
> > $\delta_l^{t+1} = \delta_l^t$, $\delta_g^{t+1} = \delta_g^t$
>
> **end**
> Until converge to **fair min-max solution** $h^1$;

**end**
**Stage 2:** continue to optimize $h^1$ to achieve Pareto optimality
**for** $t = 1, 2, ...T^2$ **do**
> *Local clients:*
> The center server sends the global model $h^t$ to all local client;
> Local clients evaluate the performance of $h^t$ and obtain the local gradients $\nabla_{\theta_t} l_i$;
> *Center server:*
> Determine $d^t$ by solving the LP problem in the Eq.(15) in the main text;
> Update the model $\theta^{t+1} = \theta^t + \eta d^t$;
> Until converge to $h^*$;

**end**
**Output:** the **fair Pareto min-max** solution $h^*$.

---

# B  Proof of Theorem 1

Firstly, $\mathcal{H}^{FP}$ and $\mathcal{H}^{FU}$ are always non-empty from their definitions. Suppose $h \in \mathcal{H}^{FU} \subset \mathcal{H}^F$. From the definitions of $\mathcal{H}^{FP}$ and $\mathcal{H}^{FU}$, we have (1) $\exists h' \in \mathcal{H}^{FP}$, s.t., $h'$ dominates $h$ that $h' \preceq h \in \mathcal{H}^{FU}$; (2) $\max(l(h)) \leq \max(l(h'))$. From (1), $\max(l(h)) \geq \max(l(h'))$ holds. Combining (2), we have $\max(l(h)) = \max(l(h'))$, so $h' \in \mathcal{H}^{FU}$ holds. Therefore, $h' \in \mathcal{H}^{FU} \cap \mathcal{H}^{FP} \neq \varnothing$ holds.

# C  Convergence Derivation of FCFL

We will prove the convergence of our method in this section. FCFL reaches a min-max Pareto fair model by a two-stage process and we aim to identify a gradient direction $d$ to optimize the model in

each iteration. For $N$ clients with $N$ corresponding objectives $[l_1, l_2, .l_N]$ and a given direction $d$, we have the following proposition, i.e.,

**Proposition 1.** *Given the gradient direction $d$ with all $d^T \nabla_\theta l_i < 0$, there exists $\eta^0$, such that:*

$$l_i(h_{\theta^{t+1}}) < l_i(h_{\theta^t}), \forall i \in \{1, 2, ., N\} \\ \theta^{t+1} = \theta^t + \eta \cdot d, \forall \eta \in [0, \eta^0]. \tag{S.1}$$

*Proof.* Consider Taylor's expansion with Peano's form of the differentiable objective function $l_i(h_{\theta^{t+1}})$:

$$l_i(h_{\theta^{t+1}}) = l_i(h_{\theta^t + \eta \cdot d}) = l_i(h_{\theta^t}) + \eta d^T \nabla_{\theta^t} l_i + o(\eta), \tag{S.2}$$

where $o(\eta)$ denotes a function that approaches 0 faster than $\eta$. Specifically, $\forall \epsilon > 0$, there exists $\eta^0$, such that $o(\eta) < \epsilon \eta$ for all $\eta \in [0, \eta^0]$. Now we set $\epsilon = -d^T \nabla_{\theta^t} l_i > 0$, there exists $\eta^0$, such that $l_i(h_{\theta^{t+1}}) - l_i(h_{\theta^t}) = \eta d^T \nabla_{\theta^t} l_i + o(\eta) = -\eta \epsilon + o(\eta) < 0$ for all $\eta \in [0, \eta^0]$. □

We will prove the convergence of our method in two steps. First, when optimizing $h$ to achieve min-max performance with MCF as in Eq.( 11) in the main text, we optimize either $\hat{l}$ or $\hat{g}'$ and keep $\hat{l}$ without ascent in each iteration. The monotonic decreasing $\hat{l}$ means $h$ will converge to $h^1$ satisfying fairness and min-max constraints. Then, we continue to optimize $h^1$ to achieve Pareto optimality by controlling the gradient direction $d$ without causing ascent for all objectives in Eq.(15) in the main text. The objective $\frac{1}{N} \sum_{i=1}^{N} l_i$ will monotonically decrease until convergence as we constrain all objectives to descend or remain unchanged defined in Eq.(15) in the main text. Assuming all objectives and their derivatives are bounded, the formal and detailed derivations are as follows.

**Convergence of Constrained Min-Max Optimization** To prove the convergence of the Constrained Min-Max Optimization procedure, we firstly give Lemma 1 as follows:

**Lemma 1.** *In each iteration, the surrogate maximum function $\hat{l}_{max}(h_{\theta^t}, \delta_l^t)$ decreases:*

$$\hat{l}_{max}(h_{\theta^{t+1}}, \delta_l^t) \leq \hat{l}_{max}(h_{\theta^t}, \delta_l^t) \tag{S.3a}$$

$$\hat{l}_{max}(h_{\theta^t}, \delta_l^{t+1}) \leq \hat{l}_{max}(h_{\theta^t}, \delta_l^t) \tag{S.3b}$$

$$\hat{l}_{max}(h_{\theta^{t+1}}, \delta_l^{t+1}) \leq \hat{l}_{max}(h_{\theta^t}, \delta_l^t). \tag{S.3c}$$

*Proof.* We prove Eq.(S.3a) in two cases: 1) $\hat{g}_{max}(h_{\theta^t}, \delta_g^t \leq 0)$: we obtain the gradient direction $d$ by solving Eq.(12) in the main text. As we choose $d = -\nabla_{\theta_t} \hat{l}$, $d^T \nabla_{\theta_t} \hat{l} < 0$ holds. According to Proposition 1, as $\min d^T \nabla_{\theta_t} \hat{l} \leq d^T \nabla_{\theta_t} \hat{l} < 0$, we prove $\hat{l}_{max}(h_{\theta^{t+1}}, \delta_l^t) \leq \hat{l}_{max}(h_{\theta^t}, \delta_l^t)$ as in Eq.(S.3a); 2) as $\hat{g}_{max}(h_{\theta^t}, \delta_g^t > 0)$, we obtain the gradient direction $d$ by solving Eq.(13) in the main text. If we choose $-d$ which lies on the angular bisector of the angle formed by $\nabla_{\theta^t} \hat{l}$ and $\nabla_{\theta^t} \hat{g}$, we have $d^T \nabla_{\theta^t} \hat{l} \leq 0$ and $d^T \nabla_{\theta^t} \hat{g} \leq 0$. As the optimal solution $d^*$ of Eq.(13) in the main text satisfies $d^{*T} \nabla_{\theta^t} \hat{l} \leq 0$ and $d^{*T} \nabla_{\theta^t} \hat{g} \leq d^T \nabla_{\theta^t} \hat{g} \leq 0$, we prove $\hat{l}_{max}(h_{\theta^{t+1}}, \delta_l^t) \leq \hat{l}_{max}(h_{\theta^t}, \delta_l^t)$ as in Eq.(S.3a) in this case. While the SMF $\hat{l}_{max}(\theta^t, \delta_l^t)$ monotonically increases with respect to $\delta_l^t$, Eq.(S.3b) holds as $\theta^{t+1} = \beta \cdot \theta^t < \theta^t$. From Eq.(S.3a) and Eq.(S.3b), we have $\hat{l}_{max}(h_{\theta^{t+1}}, \delta_l^{t+1}) \leq \hat{l}_{max}(h_{\theta^t}, \delta_l^{t+1}) \leq \hat{l}_{max}(h_{\theta^t}, \delta_l^t)$ as in Eq.(S.3c) shows. □

Combining the conclusion that $\hat{l}_{max}(h_{\theta^t}, \delta_l^t)$ decreases monotonically in each iteration in Lemma 1 with $\hat{l}_{max}(h_{\theta^t}, \delta_l^t) \geq 0$, we prove the convergence of constrained min-max optimization described in Section 4.2.

**Convergence of Constrained Pareto Optimization** Constrained Pareto optimization procedure ensures the property as follows:

**Lemma 2.** *In each iteration, the objective $l(h_{\theta^t})$ decreases or remains unchanged:*

$$l_i(h_{\theta^{t+1}}) \leq l_i(h_{\theta^t}), \ \forall i \in \{1,.,N\}. \tag{S.4}$$

*Proof.* We obtain the gradient direction $d$ by solving Eq.(15) in the main text during constrained Pareto optimization. If there exists a solution $d$ which satisfies all constraints in Eq.(15) in the main text, we have $d^T \nabla_{\theta^t} l_i \leq 0, \forall i$ which means all objectives will not increase in this iteration as Lemma 2 shows. If there is no solution to Eq.(15) in the main text, we achieve Pareto optimality and the algorithm converges. □

## D  Time Complexity Analysis

FCFL scales linearly with the dimension ($n$) of the model parameters. In our constrained min-max optimization procedure, the computation of $\hat{l}_m(h, \delta_l), \hat{g}'_m(h, \delta_g)$ in Eq.(10) has runtime of $O(N)$. With the current best LP solver [1], the LP problem with k variables and $\Omega(k)$ constraints has runtime of $O^*(k^{2.38})$ [1]. The LP problem in Eq.(12) and Eq.(12) has 2 variables and at most 4 constraints (3 constraints for $d \in \overline{G}$ and 1 constraint for $d^T \nabla_\theta \hat{l} \leq 0$) so the runtime is $O^*(2^{2.38})$. The LP problem in Eq.(15) has $N$ variables and at most $2N+2$ constraints so the runtime is $O^*(N^{2.38})$. In deep model in FL, usually $n \gg N$.

## E  Implementation Details and Additional Experimental Results

### E.1  Implementation Details

Following the experiment setting on Adult dataset in works [2, 3], we use the original Adult dataset and 66% samples are training samples and 33% are test samples. We also split the 33% as test sets and train models on the remaining 67% on eICU dataset. For the experiments in fairness constrained setting, we randomly split the eICU dataset with this ratio to run all experiments five times and report the average performance. We delete the sensitive attribute when training the model in fairness-constrained setting. All models are based on Logistic Regression and are trained and evaluated on randomly split datasets. While the original function for measuring $\Delta DP$ or $\Delta EO$ in Eq.(2) in the main text is not differentiable as there is indicator function $\hat{Y} = 1_{h(X) \geq 0.5}$, we use a surrogate differentiable function to approximate the indicator function $\hat{Y} = \frac{1}{1+(\frac{1-h(X)}{h(X)})^{10}}$ during training. We implement our method with Pytorch and determine all hyper-parameters (including learning rate, the decay rate of $\delta_l$, $\delta_g$, etc.) by evaluating different combinations on the training set. We run all experiments 5 times and report the average results with stds. For all baselines except FedAve+FairReg, we use the source implementation for comparison with the optimal hyper-parameters. For the baselines on fairness-constrained experiments given uniform fairness budgets, 1) if the baselines can satisfy MCF, we try to optimize the model disparities to achieve MCF with the optimal model utilities; 2) if the baselines cannot satisfy MCF after exhaustive trying, we minimize the model disparities with reasonable model utilities. For gradient computation, we use SGD optimizer for each local client. For more algorithm details, the source code of our method is available at https://github.com/cuis15/FCFL.

### E.2  Training Devices

We train all our models on our local Linux server with 8 GeForce RTX 2080 Ti GPUs.

### E.3  Data Asset

Adult [4] is public data that anyone can download and use it freely. eICU [5] is a dataset for which permission is required. We followed the procedure on the website https://eicu-crd.mit.edu and got the approval for this dataset.

---
[1] We use $O^*$ to hide $k^{o(1)}$ and $log^{O(1)}(1/\delta)$, $\delta$ being the relative accuracy. Detailed information is in [1]

## E.4 Additional Results on Fairness Constraint Experiment

We show the statistic results on eICU in Table 1.

Table 1: the Statistic Results on LoS Prediction with MCF.

| Methods | utility mean(%) | utility min (%) | disparity max (%) |
| --- | --- | --- | --- |
| MMCF | $65.8 \pm .0$ | $61.4 \pm .0$ | $11.9 \pm .0$ |
| FA | $69.5 \pm .05$ | $60.1 \pm .09$ | $11.5 \pm 1.2$ |
| FedAve+FairReg | $67.6 \pm .9$ | $59.3 \pm 1.2$ | $11.0 \pm 4.5$ |
| ours($\epsilon = 0.1$) | $69.1 \pm 1.3$ | **$61.4 \pm .3$** | **$9.3 \pm 1.4$** |
| ours($\epsilon = 0.05$) | $69.1 \pm 1.5$ | **$60.9 \pm .5$** | **$6.8 \pm 1.1$** |

Table 1 shows the statistical results of LoS prediction with equal fairness budgets. Our method achieves the min-max performance as we set the fairness budget $\epsilon = 0.1$. All baselines cannot reduce the disparities below 0.1. When we constrain all disparities $\Delta DP_i \leq 0.05$, our method reduces the disparities significantly compared to baselines with a comparable min-max accuracy $60.9\%$.

**Income Prediction with sensitive attribute being *gender* and equal fairness budgets**

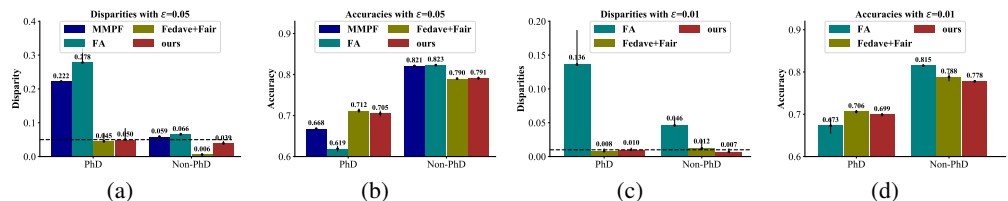

(a)  (b)  (c)  (d)

Figure 1: The disparities and accuracies on both clients as $\epsilon = 0.05$ and as $\epsilon = 0.01$ of on Adult dataset when *gender* is the sensitive attribute.

In this experiment, we select *gender* as sensitive attribute. From Figure 1, with the budget $\epsilon_i = 0.05, \forall i \in \{1,.,N\}$, we achieve comparable min-max performance with MCF while MMPF and FA violate the constraint on PhD client. As the fairness budget $\epsilon_i = 0.01, \forall i \in \{1,.,N\}$, all baselines violate the constraint on PhD client and we maintain the utilities on both clients with MCF.

**Income Prediction with sensitive attribute being *gender* and client-specific fairness budgets**

The results of Income prediction with client-specific fairness constraints in Figure 2. The disparities of both clients decrease significantly as in Figure 2(a) as $w$ decreases. With a decreasing fairness budget, the utilities of both clients slightly decreases as in Figure 2(b).

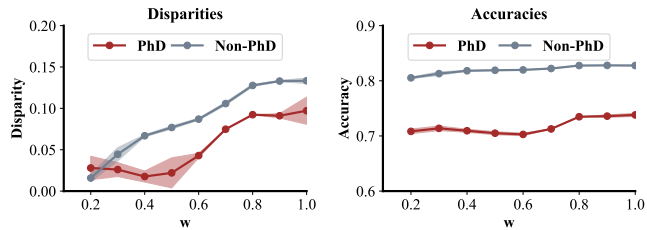

(a) The disparities vary as $w$ decreases  (b) The accuracies vary as $w$ decreases

Figure 2: Client-specific constraint experiment on Adult dataset with sensitive attribute being *gender*.

**Income prediction using EO metric** Besides DP [6] metric, we verify the effectiveness of our method using another metric *Equal opportunity* (EO) [7] which measures difference of the false negative rates across different groups as in Eq.(2). Here we show the results on Income prediction with the sensitive attribute being *race* given uniform fairness budgets $\epsilon_i = 0.05$ on both clients in our main text.

The original unconstrained model causes disparities on both clients: $\Delta EO_{PhD} = 0.105$ and $\Delta EO_{non-PhD} = 0.183$. Our model significantly reduces the disparities with fairness budget

4

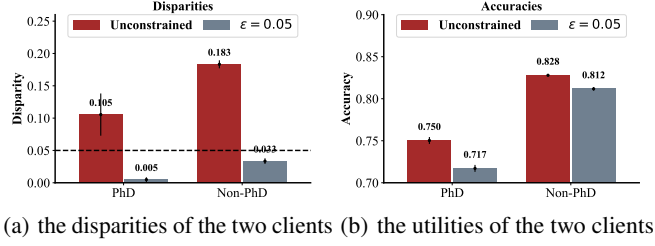

(a) the disparities of the two clients  (b) the utilities of the two clients

Figure 3: The results of unconstrained optimization and fairness-constrained optimization with the disparities measured by EO and the sensitive attribute is *race*.

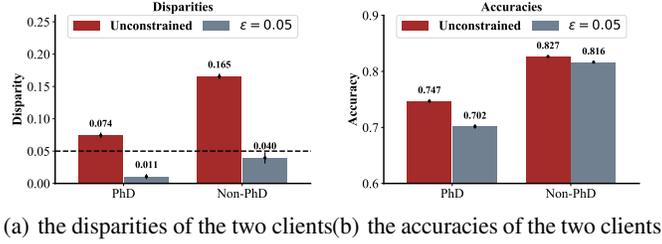

(a) the disparities of the two clients (b) the accuracies of the two clients

Figure 4: The results of unconstrained optimization and fairness-constrained optimization with the disparities measured by EO and the sensitive attribute is *gender*.

$\epsilon = 0.05$ as shown in Figure 3(a). For the model utilities on both clients shown in Figure 3(b), we achieve reasonable performances and there is a 0.03 drop on PhD client and 0.02 on non-PhD client. Similar results can also be found in Figure 4 as the sensitive attribute is *gender* in Appendix.

## F  Compatibility with Unconstrained Pareto Min-Max Optimization

### F.1  Achieving Pareto Min-Max Optimality without Fairness Constraints

Though our method is proposed for fairness-constrained multi-objective optimization, FCFL is also compatible with unconstrained min-max optimization problems which only care about the utilities of all clients as in [2, 3]:

1. optimize $h$ to achieve min-max performance:

$$\min_{d \in \overline{G}} \quad d^T \nabla_{\theta^t} \hat{l},$$

2. optimize $h$ for Pareto optimality and min-max performance:

$$\min_{d \in \overline{G}} \frac{1}{N} \sum_{i=1}^{N} d^T \nabla_{\theta^t} l_i, \quad \text{s.t. } d^T \nabla_{\theta^t} l_i \leq 0 \quad \forall i \in \{1, ., N\},$$

where $\overline{G}$ is the convex hull of $[\nabla_{\theta^t} l_1, ., \nabla_{\theta^t} l_N]$ and we obtain the gradient direction $d$ using the gradient information of all clients without accessing the local data.

### F.2  Experiments on Improving Consistency without Fairness Constraints

We evaluate the performance of our method on the problem of improving consistency without fairness constraints described with two existing methods q-FFL [2] and AFL [3].

Our method seeks for min-max performance by optimizing the SMF $\hat{l}$ which is the upper bound of all objectives. Experimental results on Income Prediction in Table 2 show that our method achieves relatively higher performance on the worst-performing client (74.9). Besides, FCFL achieves Pareto

5

Table 2: The accuracies on Income Prediction

| Methods | average (%) | PhD (%) | non-PhD (%) |
|---|---|---|---|
| q-FFL | $82.3 \pm .1$[2] | $74.4 \pm .9$[2] | $82.4 \pm .1$[2] |
| AFL | $82.5 \pm .5$[2] | $73.0 \pm 2.2$[2] | $82.6 \pm .5$[2] |
| ours | **82.7** $\pm .1$ | **74.9** $\pm .4$ | **82.8** $\pm .1$ |

optimality to avoid unnecessary harm to other clients. From Table 2, we maintain the performance on non-PhD client (82.8).

We conduct experiments on eICU dataset in unconstrained min-max setting. We randomly split the dataset 5 times and run the two prediction tasks. We show the statistical average results in Table 3. For LoS prediction task, we achieve higher uniformity by a higher utility on the worst-performing client compared to baselines. We also predict the in-hospital mortality as a prediction task and all methods achieve similar results.

Table 3: The accuracies on eICU without fairness constraints

| Methods | Mortality Prediction | | LoS Prediction | |
|---|---|---|---|---|
| | minimum (%) | average(%) | minimum (%) | average(%) |
| q-FFL | $91.7 \pm .1$ | $88.3 \pm .7$ | $57.6 \pm 2.2$ | $70.0 \pm .3$ |
| AFL | $91.7 \pm .1$ | $88.2 \pm .7$ | $58.1 \pm 2.0$ | $70.0 \pm .4$ |
| ous | $91.7 \pm .1$ | $88.3 \pm .6$ | **60.5** $\pm 2.0$ | $67.4 \pm .7$ |

## G  Discussion about Fairness Budget

### G.1  How is the fairness budget of each client relate to one another

For each fairness budegt $\epsilon_k$, it should be assigned by each client based upon its actual fairness requirements. However, the fairness budgets $\epsilon_k$ assigned by different clients are not unrelated since all fairness constraints defined by $\epsilon_k$ determine the feasible region of the model together. Due to the potential trade-off between the fairness and the utility, the fairness constraint determined by $\epsilon_k$ of the k-th client can have an impact on the utility of all clients in the federated network.

### G.2  When the fairness budget should be the same

As we stated above, the assignment of the fairness budget $\epsilon_k$ depends on the actual scenarios. In high-stake scenarios (e.g., hospitals, banks, etc.), people are highly concerned about fairness and different clients should have consistent fairness budgets. In some other scenarios such as advertising recommendations, people may be more tolerant of the difference of $\epsilon_k$ among different clients. In this case, the assignment of the fairness budget $\epsilon_k$ depends on whether it can lead to a satisfying trade-off between fairness and utility for all clients. For example, we determine the $\epsilon_k$ based upon the original disparity as the experiments presented in Sec 5.4. In addition, to obtain a budget combination that satisfies all clients, one may try different budget combinations and evaluate the performances of all clients, then all clients vote for a reasonable budget combination.

## H  Limitations and Future work

In this study, we aim to tackle the algorithmic disparity and performance inconsistency issues in federated learning. As it is hard to realize the min-max optimality and Pareto optimality by one-stage optimization, we propose a two-stage optimization framework that first achieving consistent performance then enforcing Pareto optimality. Since our framework encourages a more uniform model performance distribution, on the one hand, some clients with poor performances may be significantly improved; on the other hand, other clients whose model utility could be further improved may come to a halt.

Since we focus on addressing the local disparity, it cannot guarantee reasonable global fairness. In reality, the global disparity is also a matter of concern in a federated network. We will study how to give consideration to both local fairness and global fairness in our future work.

# References

[1] Michael B Cohen, Yin Tat Lee, and Zhao Song. Solving linear programs in the current matrix multiplication time. In *Proceedings of the 51st annual ACM SIGACT symposium on theory of computing*, pages 938–942, 2019.

[2] Tian Li, Maziar Sanjabi, Ahmad Beirami, and Virginia Smith. Fair resource allocation in federated learning. In *International Conference on Learning Representations*, 2019.

[3] Mehryar Mohri, Gary Sivek, and Ananda Theertha Suresh. Agnostic federated learning. In *International Conference on Machine Learning*, pages 4615–4625, 2019.

[4] Ron Kohavi. Scaling up the accuracy of naive-bayes classifiers: A decision-tree hybrid. In *Kdd*, volume 96, pages 202–207, 1996.

[5] Tom J Pollard, Alistair EW Johnson, Jesse D Raffa, Leo A Celi, Roger G Mark, and Omar Badawi. The eicu collaborative research database, a freely available multi-center database for critical care research. *Scientific data*, 5(1):1–13, 2018.

[6] Cynthia Dwork, Moritz Hardt, Toniann Pitassi, Omer Reingold, and Richard Zemel. Fairness through awareness. In *Proceedings of the 3rd innovations in theoretical computer science conference*, pages 214–226, 2012.

[7] Moritz Hardt, Eric Price, and Nathan Srebro. Equality of opportunity in supervised learning. In *Proceedings of the 30th International Conference on Neural Information Processing Systems*, pages 3323–3331, 2016.



\end{enumerate}
\end{document}